\documentclass{article}

\usepackage[main, final]{neurips_2026}

\usepackage[utf8]{inputenc} %
\usepackage[T1]{fontenc}    %
\usepackage{hyperref}       %
\usepackage{url}            %
\usepackage{booktabs}       %
\usepackage{amsfonts}       %
\usepackage{nicefrac}       %
\usepackage{microtype}      %
\usepackage{xcolor}
\usepackage{colortbl}
\usepackage{multirow}
\usepackage{enumitem}%
\usepackage{amsthm} %
\usepackage{amsmath} 
\usepackage{algorithm}
\usepackage{adjustbox}
\usepackage{wrapfig}
\usepackage{placeins}
\usepackage[table]{xcolor}
\usepackage{tikz}
\usepackage{arydshln}
\usepackage{algpseudocode}
\usepackage[most]{tcolorbox}

\newtheorem*{lemma*}{Lemma} 

\newtheorem*{proposition*}{Proposition}

\newtheorem{proposition}{Proposition}

\definecolor{myblue}{HTML}{3B75C3}
\definecolor{mypurple}{HTML}{6A3D9A}
\definecolor{myred}{HTML}{E33222}

\usepackage{hyperref}
\hypersetup{
    colorlinks=true,        %
    linkcolor=mypurple,     %
    citecolor=myblue,       %
    urlcolor=mypurple,      %
}
\usepackage{cleveref}
\setcitestyle{round}
\definecolor{mygreen}{HTML}{007200}
\definecolor{amber}{rgb}{1.0, 0.75, 0.0}
\definecolor{lightbluebg}{HTML}{EAF3FF}

\newcommand{\modelgroupcdashline}[1]{%
  \noalign{\vskip 2pt}%
  \cdashline{#1}%
  \noalign{\vskip 2pt}%
}
\title{From Local Skills to Long-Horizon Web Tasks: Progressive Skill Exploration for LLM Web Agents
}

\newcommand{\method}{\textsc{SCOPE}}
\DeclareRobustCommand{\crblue}[1]{{#1}}

\author{%
\crblue{Xueyang Feng\textsuperscript{1}\thanks{\crblue{Equal contribution.}}\thanks{\crblue{Work done during an internship at Huawei.}}}\quad
\crblue{Xiaohe Bo\textsuperscript{1}\footnotemark[1]}\quad
\crblue{Xu Chen\textsuperscript{1}\thanks{\crblue{Co-corresponding authors.}}}\quad
\crblue{Quanyu Dai\textsuperscript{2}\footnotemark[3]}\quad
\crblue{Feng Wen\textsuperscript{2}}\\[0.4em]
{\normalfont\small\crblue{\textsuperscript{1}Gaoling School of Artificial Intelligence, Renmin University of China}}\\
{\normalfont\small\crblue{\textsuperscript{2}Huawei Technologies Ltd.}}\\
{\normalfont\small\crblue{\texttt{xueyangfeng@ruc.edu.cn}\quad\texttt{xu.chen@ruc.edu.cn}\quad\texttt{daiquanyu@huawei.com}}}
}

\begin{document}

\maketitle

\begin{abstract}
Large language models (LLMs) are driving autonomous agents toward real-world environments, with the Web emerging as one of the most representative interaction scenarios. However, training Web agents with naive exploration remains challenging: successful trajectories are rare, and the sparse terminal rewards make it difficult for agents to obtain useful feedback from failed trajectories. Inspired by skill acquisition theory, we propose \textbf{S}kill-\textbf{Co}mposed \textbf{P}rogressive \textbf{E}xploration (\textsc{SCOPE}), a short-to-long learning framework for long-horizon Web agents that decomposes complex task exploration into a gradual expansion process from local skills to complete task-solving capability. Specifically, \textsc{SCOPE} iteratively explores around local skills: \textit{Skill Discovery} extracts reusable local skills and failure experiences from historical rollouts; \textit{Skill Orchestration} composes discovered skills into high-level skill chains; and \textit{Skill Expansion} leverages failure experiences aligned with the current skill chain to guide the agent toward exploring new subskills. Through this process, \textsc{SCOPE} progressively constructs task-completing skill chains across rollouts and internalizes them into the policy parameters through agentic on-policy distillation. Extensive experiments on WebArena, WebVoyager, and WebShop show that \textsc{SCOPE} consistently improves exploration dynamics and achieves 3.9\%--18.6\% relative gains over the strongest baselines in each setting.
\end{abstract}

\section{Introduction}
Large language models (LLMs)~\citep{qwen2025qwen25technicalreport,grattafiori2024llama3herdmodels} have demonstrated remarkable capabilities in language understanding, commonsense reasoning, and complex planning~\citep{xu2025largereasoningmodelssurvey,feng-etal-2024-large}, laying an important foundation for autonomous agents~\citep{Wang_2024}. In this context, Web agents that interact with real-world websites~\citep{zhou2024webarenarealisticwebenvironment,he2024webvoyagerbuildingendtoendweb,yao2022webshop} have become a highly promising research direction, as they hold the potential to bridge human natural language instructions and the operational workflows of traditional computer systems. Unlike conventional single-step text generation tasks, LLM-based Web agents must continuously perceive complex and dynamically changing Web environments and make long-horizon decisions throughout interaction, which poses substantially greater technical challenges.

To improve Web agents on long-horizon tasks, recent studies have begun to explore agentic reinforcement learning paradigms~\citep{qi2025webrltrainingllmweb,wei-etal-2025-webagent,feng2025groupingrouppolicyoptimizationllm}. The central idea is to enable agents to interact directly with online environments, collect trajectories, and optimize their policy parameters using task rewards. However, the naive exploration strategies used in existing agent training are often ineffective: (1) \textbf{Sparse Successes}: long-horizon interactions and dynamic webpage states make it difficult for agents to obtain successful trajectories through naive end-to-end exploration; (2) \textbf{Unexploited Failures}: rewards in Web environments are typically sparse, so failed trajectories are often discarded entirely. As a result, when successful trajectories are difficult to obtain, failed attempts provide little guidance for later exploration, often leading to repetitive trial-and-error failures. This raises a key question: \textit{how can we support effective and sustained exploration for long-horizon tasks when successful trajectories are difficult to obtain through naive exploration?}

Fitts and Posner's skill acquisition theory~\citep{fitts1967human} provides an instructive perspective for addressing these exploration challenges: complex skills are not acquired through a single end-to-end attempt, but are progressively composed from simpler skills. This perspective naturally fits long-horizon Web tasks, which typically exhibit an inherent compositional structure. Specifically, such tasks can often be decomposed into a sequence of subtasks that are locally verifiable through environmental feedback. For example, when a human first visits a shopping website, she may find it difficult to find orders with a specific status in a single attempt. However, even if the overall task fails, environmental feedback can still reveal useful sub-skills and informative failure experiences in the trajectory, such as navigating to the order page or entering irrelevant pages. In the next attempt, the former can serve as a basis for more complex skills like order sorting, while the latter can help the agent avoid repeating ineffective actions. In this way, learning a long-horizon task is decomposed into a progressive expansion process from local skills to complete task-solving capability.

Motivated by this, we propose \textbf{S}kill-\textbf{Co}mposed \textbf{P}rogressive \textbf{E}xploration (\textsc{SCOPE}), a short-to-long progressive learning framework for long-horizon Web agents. The core of \textsc{SCOPE} is an iterative exploration process organized around skills: (1) \textit{Skill Discovery}, extracting local skills and failure experiences from historical rollout trajectories; (2) \textit{Skill Orchestration}, recomposing existing skills into high-level skill chains to recover intermediate progress; and (3) \textit{Skill Expansion}, leveraging failure experiences aligned with the current skill chain to guide the agent toward exploring new subskills. Through this iterative process, \textsc{SCOPE} can progressively obtain successful trajectories by accumulating task-completing skills across different rollouts. Finally, in \textit{Skill Distillation}, we internalize the resulting task-completing skill chains into the policy parameters through agentic on-policy distillation. We evaluate \textsc{SCOPE} in three real and simulated Web environments: WebArena~\citep{zhou2024webarenarealisticwebenvironment}, WebVoyager~\citep{he2024webvoyagerbuildingendtoendweb}, and WebShop~\citep{yao2022webshop}. Across these benchmarks, \textsc{SCOPE} improves the main task metric by 3.9\%--18.6\% over the strongest corresponding baselines.

Our main contributions are as follows:

\begin{itemize}[leftmargin=*, itemsep=0pt, topsep=0pt]
\item We propose a short-to-long perspective for long-horizon Web-agent learning, decomposing Web-task exploration into a progressive process from local skills to complete task-solving capability.

\item  We propose \method{}, which enables agents to discover, organize, and expand local skills across rollouts, progressively transforming partial progress into successful trajectories. These trajectories are then internalized into the policy parameters through agentic on-policy distillation.

\item We evaluate \method{} on three real and simulated Web environments, where it consistently outperforms existing Web-agent training methods across different LLM backbones. Beyond the overall performance gains, we further analyze the source of these improvements and study how the effectiveness of two on-policy intervention strategies evolves with model capability.

\end{itemize}

\section{Background}
For a detailed survey of related work, please refer to Appendix~\ref{app:related-work}. In this section, we formalize the Web-task setting and the corresponding
Web-agent training problem that motivates our method.

\paragraph{Web-Agent Task Formulation.}
We mainly focus on Web agents that can interact with real-world websites~\citep{zhou2024webarenarealisticwebenvironment,he2024webvoyagerbuildingendtoendweb}. They access websites through a standard browser and perform browser actions based on text-based webpage observations, typically represented as HTML or AXTree. Formally, a Web task can be modeled as a long-horizon interactive decision-making problem. Given a user task \(G\), the agent interacts with the Web environment over a horizon of \(T\) steps. At each step \(t \in \{0,\dots,T-1\}\), the agent receives a webpage observation \(o_t\) and forms a state \(s_t\) together with the interaction history. The agent's policy model \(\pi_\theta\) then generates an action \(a_t \sim \pi_\theta(\cdot \mid G, s_t)\). The interaction produces a trajectory \(\tau=(s_0,a_0,s_1,a_1,\dots,s_{T-1},a_{T-1},s_T)\), which receives a sparse terminal reward \(R(\tau)\in\{0,1\}\) indicating whether the task is completed.

\paragraph{Web-Agent Training.}
To improve Web agents' ability to complete long-horizon tasks, recent work has begun training policy models by letting agents interact directly with the environment under the policy \(\pi_\theta\)~\citep{qi2025webrltrainingllmweb,he-etal-2025-openwebvoyager,zhou2025pae,yang2025sage}. Existing methods typically adopt a naive end-to-end exploration strategy: they sample a user query \(G\) from a training task distribution \(\mathcal{D}\), roll out the agent from the initial state to collect a complete trajectory, and rely on terminal rewards \(R(\tau)\) to optimize the standard objective:
\(
\max_\theta \;\; \mathbb{E}_{G\sim\mathcal{D},\,\tau\sim\pi_\theta(\cdot\mid G)}[R(\tau)].
\)

However, in long-horizon Web tasks, such exploration strategies often struggle to directly obtain complete successful trajectories. Since rewards are provided only at the terminal step, if a trajectory fails to complete the final task, it is often discarded as a whole even if some of its segments contain meaningful local progress. This motivates our key observation that intermediate webpage observations \(o_t\) record environmental changes, such as state transitions, interface changes, and error messages, and therefore provide informative feedback beyond this sparse supervision. By leveraging such feedback, we can view failed trajectories as sources of local skills and failure experiences, and decompose long-horizon task solving into subtasks that can be progressively discovered, reused, and expanded.

\section{Methodology}
\begin{figure}[t]
    \centering
    \includegraphics[width=\textwidth]{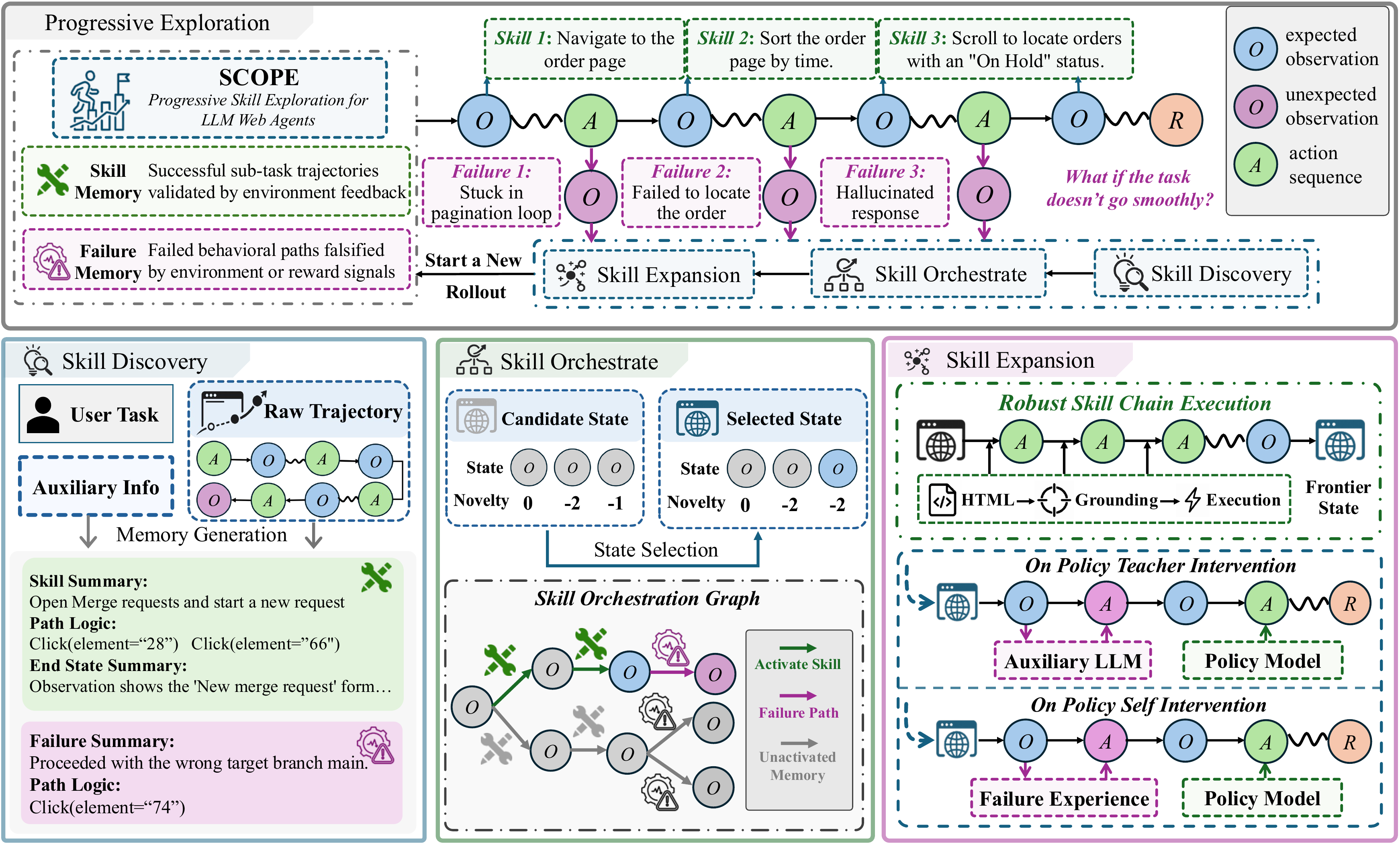}
        \vspace{-1em}
    \caption{Overview of \method{}'s progressive exploration framework: \textbf{Skill Discovery} extracts local skills and failure experiences from historical trajectories; \textbf{Skill Orchestration} composes local skills into skill chains; and \textbf{Skill Expansion} uses chain-aligned failure experiences to guide exploration.}
    \vspace{-1em}
    \label{fig:framework}
\end{figure}

Inspired by Fitts and Posner's skill acquisition theory~\citep{fitts1967human}, we propose \method{}, a short-to-long learning framework for long-horizon Web agents. As illustrated in Figure~\ref{fig:framework}, the core of \method{} is a skill-centric progressive exploration process: \textit{Skill Discovery} identifies reusable local skills and failure experiences from historical trajectories (Section~\ref{sec:discovery}); \textit{Skill Orchestration} composes validated skills into skill chains (Section~\ref{sec:composition}); and \textit{Skill Expansion} uses chain-aligned failures to guide exploration beyond the current frontier (Section~\ref{sec:guided_rollout}). Finally, the resulting task-completing skill chains are internalized into the policy through agentic on-policy distillation (Section~\ref{sec:distillation}).

\subsection{Skill Discovery}
\label{sec:discovery}

Skill acquisition theory suggests that, in the early stage of learning, learners facing complex long-horizon tasks often rely on explicit task decomposition to gradually build an understanding of the task structure. Following this view, we identify local skills and failure experiences from rollout trajectories. 
Specifically, given a rollout trajectory \(\tau\), we use the task goal \(G\), the task-completion criterion \(c\), and the intermediate environmental feedback contained in \(\tau\) to parse it into semantically contiguous subtask units and assess whether each unit achieves its intended subgoal:
\[
(\xi_i, y_i)_{i=1}^n = \textsc{Discover}(G, c, \tau), \quad
\xi_i=(s_{\texttt{start}}^{(i)}, A^{(i)}, s_{\texttt{end}}^{(i)}; m_i), \; y_i \in \{0,1\}.
\]
Here, \(\textsc{Discover}\) is implemented by an auxiliary LLM with a task-specific prompt, as are the symbolic operators introduced below. The criterion \(c\) typically includes goal-related constraints or target page states that help assess whether a subgoal has been achieved. In each segment, \(\texttt{start}\) and \(\texttt{end}\) refer to its boundaries in the original trajectory, and \(s_{\texttt{start}}^{(i)}\) and \(s_{\texttt{end}}^{(i)}\) denote the corresponding starting and ending states. \(A^{(i)}=(a_{\texttt{start}}^{(i)},\dots,a_{\texttt{end}-1}^{(i)})\) denotes the action subsequence that transitions between them, \(m_i\) denotes the semantic subtask description, and \(y_i\) indicates whether the intended subgoal is achieved. When \(y_i=1\), the corresponding segment represents a local action sequence in the original trajectory that successfully completes a specific Web subtask; to enable reuse in subsequent rollouts, we maintain it as a skill. When \(y_i=0\), the corresponding segment represents a behavioral sequence that attempts but fails to complete a specific subtask, and we store it as failure experience. In implementation, we store non-redundant entries in two memory banks:
\[
\mathcal{B}_{\mathrm{skill}} \leftarrow \mathcal{B}_{\mathrm{skill}} \cup \{\xi_i \mid y_i=1\}, \quad
\mathcal{B}_{\mathrm{fail}} \leftarrow \mathcal{B}_{\mathrm{fail}} \cup \{\xi_i \mid y_i=0\}.
\]
These two memory banks capture complementary intermediate feedback. The skill memory \(\mathcal{B}_{\mathrm{skill}}\) stores subtask-completion sequences verified by environmental feedback, each capturing a reusable transition between environment states. Notably, such progress does not necessarily lead to final task success. The failure memory \(\mathcal{B}_{\mathrm{fail}}\) records failure patterns triggered along particular progress paths. In the following sections, we study how to use these memories to support subsequent exploration.

\subsection{Skill Orchestration}
\label{sec:composition}

In the middle stage of skill acquisition, a learner composes previously discovered local rules into higher-level complex skills. Inspired by this view, we compose validated skills in the skill bank \(\mathcal{B}_{\mathrm{skill}}\) into reusable skill chains for subsequent attempts. Concretely, the skill bank can be viewed as a state graph connected by local skills, where each node \(s_k\) denotes a specific environment state, such as an order page or a sorting page, and each edge corresponds to a validated local skill. We collect candidate reachable nodes in this graph as the frontier state set \(\mathcal{S}_{\mathrm{frontier}}=\{s_k\}_{k=1}^{K}\), and perform group-level evaluation to select the frontier state \(s^\star = \textsc{Select}\!\left(G,\, c,\, \mathcal{S}_{\mathrm{frontier}}\right)\) for the next exploration rollout.
After selecting \(s^\star\), we trace and compose a skill chain \(p^\star\) of connected subskills from the initial state \(s_0\) to \(s^\star\). In the next rollout, this skill chain is invoked to restore the environment to the frontier state \(s^\star\), which serves as the basis for further exploration. 

In this process, we use two implementation details. \textit{Exploration-aware frontier selection}: we maintain a visitation count for each frontier state and prompt the LLM to downweight frequently visited states, balancing progress reuse and exploration diversity~\citep{ecoffet2021firstreturn,lu2025intelligentgoexplore}. \textit{Robust skill execution}: Original agent trajectories often refer to target elements by indices or positions in AXTree/HTML. To make skills reusable across dynamic webpages, we extract element names from the original identifiers and ground them to the current AXTree/HTML context before execution.

\subsection{Skill Expansion}
\label{sec:guided_rollout}

Once skills have been organized into a chain, failure experiences aligned with it can guide exploration beyond the current skill graph. Specifically, we identify failure experiences in \(\mathcal{B}_{\mathrm{fail}}\) whose starting state matches the selected frontier state \(s^\star\):
\(
\mathcal{F}^\star =
\{\,\xi_i \in \mathcal{B}_{\mathrm{fail}} \mid s_{\texttt{start}}^{(i)}=s^\star\,\}.
\)
The key question is then how to effectively inject failure experiences into subsequent exploration. To this end, we study two on-policy intervention strategies applied at the first frontier action:

\begin{itemize}[leftmargin=*, itemsep=0pt, topsep=0pt]
    \item \textit{On-policy Teacher Intervention} (\textit{OPTI}). Inspired by on-policy distillation and imitation learning~\citep{agarwal2024onpolicydistillationlanguagemodels,ross2011reductionimitationlearningstructured}, we introduce a teacher model to directly generate an intervening action from \(s^\star\) and \(\mathcal{F}^\star\):
\(
a^{\mathrm{teach}} = \textsc{Intervene}(G,\, s^\star,\, \mathcal{F}^\star).
\)
This action replaces the policy model's original action at the frontier state, providing stronger external guidance.
    \item \textit{On-policy Self Intervention} (\textit{OPSI}). Inspired by on-policy self-distillation~\citep{zhao2026selfdistilledreasoneronpolicyselfdistillation}, the policy conditions on retrieved failure experiences to generate the intervening action:
\(
a^{\mathrm{self}} \sim \pi_\theta(\cdot \mid G,\, s^\star,\, \textsc{Summarize}(G,\, p^\star,\, \mathcal{F}^\star)).
\)
This variant lets the policy use failure experiences to correct its own behavior, thereby generating intervening actions closer to the policy model itself.
\end{itemize}

Importantly, these interventions remain on-policy at the level of agent-environment interaction steps: the frontier state \(s^\star\) is reached by skill-chain-based state restoration, which composes locally verified skills produced by the policy model in historical rollouts.
After the intervention, the policy model continues subsequent exploration autonomously from the updated state. We further analyze the effects and differences between these two intervention strategies in Section~\ref{sec:intervention-dynamics}.  %

\subsection{Skill Distillation}
\label{sec:distillation}
At the final stage of skill acquisition, learners internalize acquired capabilities into stable and efficient task execution. Accordingly, we adopt an agentic on-policy distillation strategy to internalize the high-quality behaviors induced by progressive exploration and memory-guided intervention into the model parameters. Let \(q_{\mathrm{SCOPE}}(\tilde{\tau}\mid G)\) denote the teacher trajectory distribution produced by \method{}'s progressive exploration. Each \(\tilde{\tau}\) is a task-completing teacher trajectory with \(R(\tilde{\tau})=1\), composed from skill chains accumulated across multiple rollouts. We distill this teacher trajectory distribution constructed by progressive exploration into the policy model \(\pi_\theta\) by minimizing
{\setlength{\abovedisplayskip}{4pt}%
\setlength{\belowdisplayskip}{4pt}%
\[
\mathcal{L}_{\mathrm{distill}}(\theta)
=
\mathbb{E}_{G\sim\mathcal{D},\,\tilde{\tau}\sim q_{\mathrm{SCOPE}}(\cdot\mid G)}
[
-\sum_{t=0}^{|\tilde{\tau}|-1}
\log \pi_\theta(\tilde{a}_t \mid G, \tilde{s}_t)
].
\]
}
Here, \((\tilde{s}_t,\tilde{a}_t)\) are the state-action pairs actually executed in the teacher trajectory constructed by progressive exploration. The teacher trajectory mixes actions from two sources: non-intervention actions sampled from the policy itself, and frontier intervention actions generated with failure-experience guidance or auxiliary-model assistance. Taking \textit{OPSI} as an example, frontier actions are produced from the failure-conditioned distribution \(\pi_\theta(a\mid G,\tilde{s}_t,\mathcal{F}^\star)\), while distillation compresses them into the standard policy distribution \(\pi_\theta(a\mid G,\tilde{s}_t)\). This process internalizes task-completing skill chains that rely on progressive exploration and failure correction into the student policy. Moreover, beyond the exploration perspective, the overall advantages of \method{} can also be understood from the perspective of on-policy distillation. We provide a detailed discussion in Appendix~\ref{app:distillation-view}.

\subsection{Overall Progressive Learning Process}
\label{sec:overall-process}
Overall, \method{} is a skill-organized framework that interleaves progressive exploration with skill distillation. From the current policy \(\pi_\theta^{(r)}\), each training round performs an initial rollout and then runs iterative exploration, where \textit{Skill Discovery}, \textit{Skill Orchestration}, and \textit{Skill Expansion} extend the frontier from local skills toward task-completing skill chains. \textit{Skill Distillation} internalizes the resulting chains into the policy, producing \(\pi_\theta^{(r+1)}\) for the next round:
{\small
\[
\pi_\theta^{(r)}
\!\rightarrow\!
\tau_0^{(r)}
\!\rightarrow\!
\underbrace{
\textsc{Discovery}\!\rightarrow\!\textsc{Orchestration}\!\rightarrow\!\textsc{Expansion}
}_{\text{Iteration }j}
\!\rightarrow\!
\tau_j^{(r)}
\!\rightarrow\!\cdots\!\rightarrow\!
\tau_J^{(r)}
\!\xrightarrow{\textsc{Distill.}}\!
\pi_\theta^{(r+1)} .
\]
}
Here, \(r\) indexes the outer training round, and \(j=1,\dots,J\) indexes its progressive exploration iterations before skill distillation. The updated policy then enters the next round, allowing subsequent exploration to benefit from accumulated memories and capabilities internalized into the model parameters. We provide the complete algorithm in Appendix~\ref{app:algorithm} to clarify \method{}'s training process.

We analyze \method{}'s progressive exploration mechanism from a variance perspective, showing that it can reduce the instability of rollouts required for long-horizon exploration from exponential to polynomial scale, thereby enabling stable and sustained exploration on long-horizon tasks.

\begin{proposition}[Polynomial Variability of Progressive Exploration]
Assume that making a locally correct frontier-expanding decision has probability at least $p\in(0,1]$, and that, at any fixed depth, the number of webpage functional patterns relevant to the current subtask progression is bounded. Then the standard deviation of exploration rollouts \(T_{\mathrm{total}}\) satisfies
\(
\mathrm{Std}(T_{\mathrm{total}})
=
O\!\left(\frac{L^{(1+2\alpha)/2}}{p}\right).
\)
\end{proposition}

Here \(L\) denotes the target task depth, and \(\alpha\) is a task-dependent constant representing the key constraint factors that must be progressively satisfied to complete the Web task. By contrast, under naive direct exploration, completing a task of depth \(L\) requires making all \(L\) critical decisions correctly, so the success probability decays as \(p^L\). This causes both the waiting time and its variability to grow exponentially with task depth. The full derivation is deferred to Appendix~\ref{app:theory}.

\section{Experiments}
To comprehensively validate the effectiveness of our proposals, we conduct extensive experiments, which are outlined as follows:
\begin{itemize}[leftmargin=*, label=\(\circ\), itemsep=0pt, topsep=0pt]
    \item First, we use \method{} to train two backbone models on both offline and online Web benchmarks, and compare them with a range of baselines to validate the benefit of progressive exploration for improving model capability and its broad applicability (Section~\ref{sec:main-results}).
    \item Second, we analyze \method{}'s progressive exploration dynamics across task depths, tracing how improved exploration translates into stronger long-horizon capability (Section~\ref{sec:exploration-dynamics}).
    \item Third, we compare the two intervention strategies, \textsc{OPTI} and \textsc{OPSI}, and analyze how their exploration and internalization behaviors vary with model capability (Section~\ref{sec:intervention-dynamics}).
    \item Finally, we further study \method{}'s scalability, self-managed memory, component necessity, and training efficiency (Section~\ref{sec:more-analysis}).
\end{itemize}

\subsection{Experiments Setup}
\paragraph{Benchmarks and Baselines.}
We evaluate \method{} on three representative Web-agent benchmarks with diverse interaction settings:
(1) \textbf{\textsc{WebArena-Lite}}~\citep{liu2024visualagentbench} is an offline hosted Web environment with long-horizon tasks across Reddit, Shopping, CMS, and GitLab, involving complex information acquisition and page-state manipulation. We follow the experimental setup of \textsc{WebRL}~\citep{qi2025webrltrainingllmweb} and use its standard leaderboard results as the main comparison baseline;
(2) \textbf{\textsc{WebVoyager}}~\citep{he2024webvoyagerbuildingendtoendweb} evaluates agents through interactions with real online websites, such as GitHub and ArXiv, where the interaction process is more dynamic and non-stationary. We use the tasks provided by \textsc{OpenWebVoyager}~\citep{he-etal-2025-openwebvoyager} as the training set and the original \textsc{WebVoyager} tasks as the test set. Since online Web environments are generally difficult to accommodate purely online reinforcement learning methods~\citep{qi2025webrltrainingllmweb}, we mainly compare with \textsc{SFT} and \textsc{RFT};
(3) \textbf{\textsc{WebShop}}~\citep{yao2022webshop} is a benchmark widely used for evaluating reinforcement-learning algorithms for Web agents. Since it supports repeatable sampling and has lower interaction cost, we further compare with online RL baselines such as \textsc{GRPO}~\citep{shao2024deepseekmathpushinglimitsmathematical}.
More details about the benchmarks and baselines are provided in Appendix~\ref{app:experimental-details}.

\paragraph{Training Details.}
We use Qwen2.5-3B~\citep{qwen2025qwen25technicalreport} and Llama3.1-8B~\citep{grattafiori2024llama3herdmodels} as policy backbones throughout our experiments. Before \method{} training, each policy undergoes a benchmark-appropriate warm-up stage; details are provided in Appendix~\ref{app:experimental-details}. For memory management, we consider both external-model and self-managed settings. In the external-model setting, the auxiliary functions \textsc{Discover}, \textsc{Select}, \textsc{Summarize}, and \textsc{Intervene} are implemented with DeepSeek-V3.2~\citep{deepseekai2025deepseekv32pushingfrontieropen}. In the self-managed setting, these functions are performed by a Llama3.1-8B model fine-tuned on pre-collected memory-management data. Prompts for all auxiliary-model calls are provided in Appendix~\ref{app:experimental-details}. For progressive exploration, we run 5 inner exploration iterations per outer training round on \textsc{WebArena} and \textsc{WebVoyager}, and 4 iterations on \textsc{WebShop}. Training is conducted with the LLaMA-Factory framework~\citep{zheng2024llamafactory}. The policy learning rate is selected from \(\{10^{-5}, 10^{-6}\}\), the batch size is 32, and the number of training epochs is 3. Unless otherwise specified, we report \textsc{OPSI} results after one outer training round; results with three outer training rounds are analyzed in Section~\ref{sec:more-analysis}.

\paragraph{Test Details.}
For environment interaction, real benchmarks are executed in browser environments. \textsc{WebVoyager} provides AXTree-based observations, while \textsc{WebArena} provides HTML-based observations. We use vLLM~\citep{kwon2023pagedattention} for policy inference. To ensure reproducibility, we set the decoding temperature to 0 during evaluation. The maximum number of interaction steps is 10 for \textsc{WebShop}, 15 for \textsc{WebVoyager}, and 30 for \textsc{WebArena}. Because some website tasks are unreachable, we filter out a subset of websites in \textsc{WebArena} and \textsc{WebVoyager}; details are provided in Appendix~\ref{app:experimental-details}. All prompts are kept consistent with the corresponding baselines.

\subsection{Main Results}
\label{sec:main-results}

\begin{table}[t!]
\centering
\caption{Results of \method{} on \textsc{WebArena}, an offline yet realistic Web benchmark. \textsc{OPTI} and \textsc{OPSI} denote on-policy Teacher Intervention and on-policy Self Intervention, respectively.}
\vspace{-1mm}
\label{tab:webarena-results}
\begin{adjustbox}{width=0.85\textwidth, center} 
\small 
\setlength{\tabcolsep}{4pt}
\begin{tabular}{llccccc}
\toprule
\textbf{Model} & \textbf{Method} & \textbf{Total SR} & \textbf{Reddit} & \textbf{GitLab} & \textbf{Shopping} & \textbf{CMS} \\
\midrule
\multicolumn{7}{c}{\it Prompting Method} \\
Qwen2.5-3B & - & 7.0 & 5.3 & 13.3 & 4.4 & 5.7 \\
Llama3.1-8B & - & 7.8 & 5.3 & 10.0 & 8.9 & 5.7 \\
OAI-4o & -  & 13.2 & 10.5 & 10.0 & 11.1 & 20.0 \\
\midrule
\multicolumn{7}{c}{\it Finetuning Method} \\

\multirow{4}{*}{Qwen2.5-3B} &  Behavior Cloning & 20.2 & 42.1 & 16.7 & 11.1 & 22.9 \\
 & \textsc{RFT}~\citep{yuan2023scalingrelationshiplearningmathematical} & 36.5 & 47.4 & 40.0 & 17.8 & 48.6 \\
 & \cellcolor{lightbluebg} \method\ (\textsc{OPTI}) & \cellcolor{lightbluebg} 41.6 \scriptsize{\textcolor{red}{(+14.0\%)}} & \cellcolor{lightbluebg} 73.7 & \cellcolor{lightbluebg} 44.4 & \cellcolor{lightbluebg} 18.2 & \cellcolor{lightbluebg} 51.4 \\
 & \cellcolor{lightbluebg} \method\ (\textsc{OPSI}) & \cellcolor{lightbluebg} 40.2 \scriptsize{\textcolor{red}{(+10.1\%)}} & \cellcolor{lightbluebg} 52.6 & \cellcolor{lightbluebg} 51.7 & \cellcolor{lightbluebg} 22.7 & \cellcolor{lightbluebg} 45.7 \\
\modelgroupcdashline{1-7}
\multirow{7}{*}{Llama3.1-8B} & Behavior Cloning & 14.9 & 6.0 & 15.0 & 14.0 & 11.0 \\
& \textsc{AWR}~\citep{peng2019advantageweightedregressionsimplescalable} & 29.5 & 57.9 & 26.7 & 17.8 & 31.4 \\
& \textsc{DigiRL}~\citep{bai2024digirltraininginthewilddevicecontrol} & 31.0 & 57.9 & 26.7 & 17.8 & 37.1 \\
 & \textsc{RFT}~\citep{yuan2023scalingrelationshiplearningmathematical} & 40.9 & \underline{68.4} & 46.7 & 26.7 & 37.1 \\
 & \textsc{WebRL}~\citep{qi2025webrltrainingllmweb} & 45.8 & 63.2 & 46.7 & 31.1 & \underline{54.3} \\
& \cellcolor{lightbluebg} \method\ (\textsc{OPTI}) & \cellcolor{lightbluebg} \underline{53.6} \scriptsize{\textcolor{red}{(+17.0\%)}} & \cellcolor{lightbluebg} \textbf{84.2} & \cellcolor{lightbluebg} \textbf{61.4} & \cellcolor{lightbluebg} \underline{40.0} & \cellcolor{lightbluebg} 51.4 \\
& \cellcolor{lightbluebg} \method\ (\textsc{OPSI}) & \cellcolor{lightbluebg} \textbf{54.3} \scriptsize{\textcolor{red}{(+18.6\%)}} & \cellcolor{lightbluebg} \textbf{84.2} & \cellcolor{lightbluebg} \underline{50.0} & \cellcolor{lightbluebg} \textbf{42.2} & \cellcolor{lightbluebg} \textbf{57.1} \\
\bottomrule
\end{tabular}
\end{adjustbox}
\end{table}

\begin{table}[t!]
\centering
\caption{Results of \method{} on \textsc{WebVoyager}, which evaluates agents on real online websites. }%
\vspace{-1mm}
\label{tab:webvoyager-results}
\begin{adjustbox}{width=0.99\textwidth, center}
\setlength{\tabcolsep}{2.5pt}
\begin{tabular}{llccccccccc}
\toprule
\textbf{Model} & \textbf{Method} & \textbf{Total SR} & \textbf{Amazon} & \textbf{Apple} & \textbf{ArXiv} & \textbf{BBC} & \textbf{Cambridge} & \textbf{GitHub} & \textbf{HFace} & \textbf{Wolfram} \\
\midrule
\multicolumn{11}{c}{\it Prompting Method} \\
OAI-4o & - & 42.2 & 51.3 & 41.9 & 27.9 & 32.6 & 41.9 & 36.6 & 46.5 & 56.5 \\
DeepSeek-v3.2 & - & 42.0 & 46.2 & 40.6 & 34.6 & 40.6 & 29.0 & 45.0 & 20.7 & 65.2 \\
\midrule
\multicolumn{11}{c}{\it Finetuning Method} \\
\multirow{3}{*}{Qwen2.5-3B} & \textsc{SFT} & 24.1 & 13.2 & \textbf{31.2} & 19.0 & 9.4 & \textbf{14.0} & 34.1 & 27.0 & 41.3 \\
 & \textsc{RFT} & 26.1 & 18.4 & 25.0 & 21.4 & \textbf{21.9} & 0.0 & \textbf{46.3} & 21.6 & \textbf{51.1} \\
& \cellcolor{lightbluebg} \method & \cellcolor{lightbluebg} \textbf{28.4} \scriptsize{\textcolor{red}{(+8.8\%)}} & \cellcolor{lightbluebg} \textbf{23.7} & \cellcolor{lightbluebg} \textbf{31.2} & \cellcolor{lightbluebg} \textbf{38.1} & \cellcolor{lightbluebg} 9.4 & \cellcolor{lightbluebg} 7.1 & \cellcolor{lightbluebg} 34.1 & \cellcolor{lightbluebg} \textbf{32.4} & \cellcolor{lightbluebg} 45.7 \\
\modelgroupcdashline{1-11}
\multirow{3}{*}{Llama3.1-8B} & \textsc{SFT} & 33.4 & 18.4 & 25.0 & 38.1 & 9.4 & \textbf{16.3} & 58.5 & \textbf{35.1} & 52.2 \\
 & \textsc{RFT} & 37.3 & 42.1 & 34.4 & \textbf{52.4} & 15.6 & 9.3 & 48.8 & 32.4 & \textbf{56.5} \\
& \cellcolor{lightbluebg} \method & \cellcolor{lightbluebg} \textbf{42.8} \scriptsize{\textcolor{red}{(+14.7\%)}} & \cellcolor{lightbluebg} \textbf{52.6} & \cellcolor{lightbluebg} \textbf{37.5} & \cellcolor{lightbluebg} \textbf{52.4} & \cellcolor{lightbluebg} \textbf{28.1} & \cellcolor{lightbluebg} \textbf{16.3} & \cellcolor{lightbluebg} \textbf{65.9} & \cellcolor{lightbluebg} \textbf{35.1} & \cellcolor{lightbluebg} 50.0 \\
\bottomrule
\end{tabular}
\vspace{-1.25em}
\end{adjustbox}
\end{table}

\textbf{\method{} achieves stronger overall performance than existing baselines.}
Table~\ref{tab:webarena-results}, Table~\ref{tab:webvoyager-results}, and Table~\ref{tab:webshop-results} summarize the main comparisons. Overall, general models without task-specific training perform substantially worse on complex interactive Web tasks than fine-tuned models, highlighting the importance of fine-tuning Web agents. 
Moreover, methods that optimize the policy through environment interaction, such as \textsc{RFT} and \textsc{GRPO}, generally outperform \textsc{SFT}, suggesting that direct interaction and reward-based improvement provide an effective path for enhancing Web-agent capabilities. Beyond these methods, \method{} adopts a short-to-long progressive exploration strategy and uses agentic on-policy distillation to internalize effective skills discovered across rollouts. 
\begin{wraptable}{r}{0.5\textwidth}
\vspace{-0.6em}
\centering
\caption{\method{} outperforms online RL methods in the simulated \textsc{WebShop} environment.}
  \vspace{0.4em}
\label{tab:webshop-results}
\begin{adjustbox}{width=\linewidth}
\scriptsize
\setlength{\tabcolsep}{3pt}
\begin{tabular}{llcc}
\toprule
\textbf{Model} & \textbf{Method} & \textbf{Score} & \textbf{Succ. SR} \\
\midrule
\multirow{4}{*}{Qwen2.5-3B}
& \textsc{SFT} & 0.731 & 55.6 \\
& \textsc{RFT} & 0.780 & 60.0 \\
& \textsc{GRPO} & 0.898 & 80.8 \\
& \cellcolor{lightbluebg}\method
& \cellcolor{lightbluebg}\textbf{0.933} \scriptsize{\textcolor{myred}{(+3.9\%)}}
& \cellcolor{lightbluebg}\textbf{84.6} \scriptsize{\textcolor{myred}{(+4.7\%)}} \\
\modelgroupcdashline{1-4}
\multirow{4}{*}{Llama3.1-8B}
& \textsc{SFT} & 0.719 & 52.6 \\
& \textsc{RFT} & 0.833 & 68.0 \\
& \textsc{GRPO} & 0.889 & 76.2 \\
& \cellcolor{lightbluebg}\method
& \cellcolor{lightbluebg}\textbf{0.938} \scriptsize{\textcolor{myred}{(+5.5\%)}}
& \cellcolor{lightbluebg}\textbf{85.2} \scriptsize{\textcolor{myred}{(+11.8\%)}} \\
\bottomrule
\end{tabular}
\vspace{-1.25em}
\end{adjustbox}
\end{wraptable}

As a result, \method{} consistently outperforms the strongest corresponding baselines across \textsc{WebArena}, \textsc{WebVoyager}, and \textsc{WebShop}, achieving relative gains of 3.9\%--18.6\%.

\textbf{\method{} shows broad applicability across Web environments and model scales.}
(1) \textbf{Environment perspective}: \textsc{WebArena} and \textsc{WebVoyager} go beyond low-cost simulators and involve realistic or online Web environments, where interactions are often non-stationary, expensive, and difficult to parallelize. These properties make standard agentic RL methods, such as \textsc{GRPO}, difficult to apply effectively in such settings. By contrast, \method{} consistently improves performance in both realistic online and controllable simulated environments, suggesting that its local-skill-driven exploration mechanism remains applicable in dynamic Web environments.
(2) \textbf{Model perspective}: As shown in Table~\ref{tab:webarena-results}, Table~\ref{tab:webvoyager-results}, and Table~\ref{tab:webshop-results}, \method{} improves both Qwen2.5-3B and Llama3.1-8B, showing that it is not limited to a particular model scale. Moreover, Llama3.1-8B obtains more pronounced gains than Qwen2.5-3B, suggesting that \method{}'s progressive exploration can scale with model capability.

\subsection{Tracing Exploration Dynamics to Long-Horizon Capability}
\label{sec:exploration-dynamics}
\begin{figure}[t!]
\centering
\includegraphics[width=\textwidth]{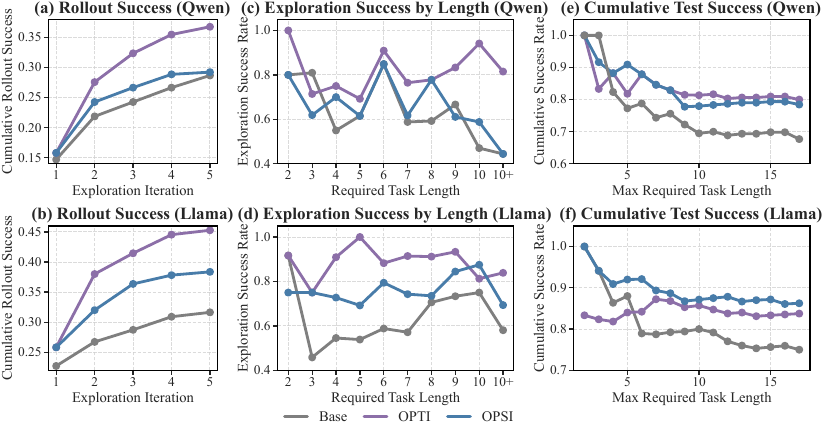}
\vspace{-2em}
\caption{Long-horizon exploration dynamics and intervention behavior of \method{} during training. Rows correspond to Qwen2.5-3B and Llama3.1-8B; columns show cumulative rollout success, exploration success by task length, and cumulative test success by maximum task length.}
\label{fig:model-analysis}
\vspace{-1.25em}
\end{figure}

To further understand the source of \method{}'s performance gains, we conduct a systematic mechanistic analysis of its progressive exploration process, tracing how improved exploration dynamics translate into stronger long-horizon model capability.
(1) \textbf{\method{} exhibits more stable and sustained exploration ability}. As shown in Figure~\ref{fig:model-analysis}(a,b), for the same number of rollouts, \method{} discovers more reward-positive samples than direct rollouts, indicating more effective exploration.
(2) \textbf{\method{} shows a clear exploration advantage on longer and harder tasks.} We group successful rollouts by the required number of steps for each task. We exclude tasks that cannot be solved by any model, and use the shortest successful trajectory length among all models as the estimated required length for each remaining task. As shown in Figure~\ref{fig:model-analysis}(c,d), \method{} achieves higher rollout success rates on longer tasks, suggesting that its progressive exploration mechanism helps reach task states that are difficult to access through direct rollouts.
(3) \textbf{The model trained with \method{} obtains stronger performance gains on long-horizon tasks}. As shown in Figure~\ref{fig:model-analysis}(e,f), \method{}-trained models maintain higher cumulative test success on tasks that require longer interaction trajectories, further demonstrating the clear advantage of \method{}-trained models on long-horizon tasks.

\FloatBarrier

\subsection{Understanding the Effectiveness of \textsc{OPTI} and \textsc{OPSI}}
\label{sec:intervention-dynamics}
Table~\ref{tab:webarena-results} shows that both \textsc{OPTI} and \textsc{OPSI} improve WebArena performance, but their relative advantages depend on model capability. We explain this pattern from two aspects:
(1) \textbf{Exploration.} As shown in Figure~\ref{fig:model-analysis}(c,d), \textsc{OPTI} achieves stronger rollout performance than \textsc{OPSI}, because auxiliary-model action intervention can more directly push the agent beyond its current exploration frontier. This is especially helpful for weaker policies such as Qwen2.5-3B.
(2) \textbf{Internalization.} Although \textsc{OPSI} is less aggressive than \textsc{OPTI} during exploration, its corrective actions are generated by the policy model itself. For stronger policies such as Llama3.1-8B, these self-corrected behaviors are closer to the model's own policy distribution and are therefore easier to internalize through distillation. This observation is consistent with prior findings on student-friendly distillation~\citep{yuan2023studentfriendlyknowledgedistillation}.

\subsection{Further Analysis of Scalability, Self-Managed Memory, Ablations, and Efficiency}
\label{sec:more-analysis}

\begin{figure}[t!]
\centering
\newcommand{\moreanalysispaneltitle}[1]{\centerline{\fontsize{8.4}{9.4}\selectfont\bfseries #1}}
\includegraphics[width=0.96\textwidth]{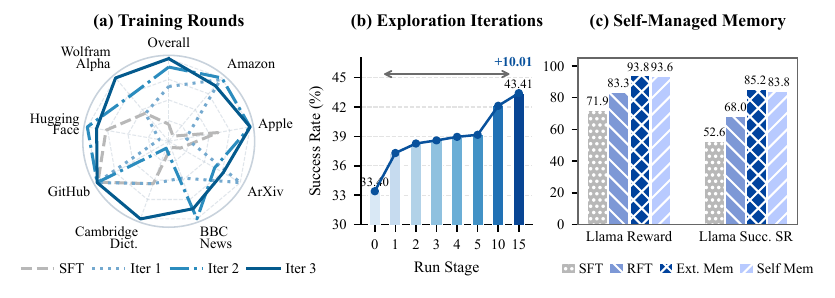}
\vspace{-1.25em}

\begin{tabular}{@{}c@{\hspace{0.012\textwidth}}c@{}}
\begin{minipage}[t]{0.48\textwidth}
\centering
\vspace{0pt}
\moreanalysispaneltitle{(d) Ablation Study}
\vspace{0.28em}
\footnotesize
\setlength{\tabcolsep}{2.5pt}
\renewcommand{\arraystretch}{1.08}
\newcommand{\moreanalysisdelta}[1]{{\fontsize{5.0}{5.6}\selectfont #1}}
\begin{tabular}{lcc}
\toprule
\textbf{Method} & \textbf{Reward} & \textbf{Succ. SR} \\
\midrule
\cellcolor{lightbluebg}\method{}
& \cellcolor{lightbluebg}\textbf{0.933}
& \cellcolor{lightbluebg}\textbf{84.6} \\
\midrule
w/o Skill Discovery & 0.827 \moreanalysisdelta{\textcolor{mygreen}{(-11.4\%)}} & 66.6 \moreanalysisdelta{\textcolor{mygreen}{(-21.3\%)}} \\
w/o Skill Orchestration & 0.897 \moreanalysisdelta{\textcolor{mygreen}{(-3.9\%)}} & 77.0 \moreanalysisdelta{\textcolor{mygreen}{(-9.0\%)}} \\
w/o Skill Expansion & 0.816 \moreanalysisdelta{\textcolor{mygreen}{(-12.5\%)}} & 67.2 \moreanalysisdelta{\textcolor{mygreen}{(-20.6\%)}} \\
\midrule
\textsc{RFT} & 0.780 \moreanalysisdelta{\textcolor{mygreen}{(-16.4\%)}} & 60.0 \moreanalysisdelta{\textcolor{mygreen}{(-29.1\%)}} \\
\textsc{SFT} & 0.731 \moreanalysisdelta{\textcolor{mygreen}{(-21.7\%)}} & 55.6 \moreanalysisdelta{\textcolor{mygreen}{(-34.3\%)}} \\
\bottomrule
\end{tabular}
\renewcommand{\arraystretch}{1.0}
\end{minipage}
&
\begin{minipage}[t]{0.47\textwidth}
\centering
\vspace{0pt}
\moreanalysispaneltitle{(e) Efficiency Frontier}
\vspace{0.28em}
\includegraphics[width=0.96\linewidth]{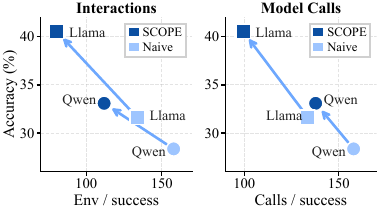}
\end{minipage}
\end{tabular}
\vspace{-0.35em}
\caption{Further analysis of \method{}. (a) \method{} scales with outer training rounds. (b) \method{} scales with progressive exploration iterations. (c) Self-managed memory achieves performance comparable to external memory. (d) Key components of \method{} contribute to final performance. (e) \method{} achieves a better training-efficiency trade-off in environment interactions and model calls.}
\vspace{-1.25em}
\label{fig:more-analysis}
\end{figure}

\textbf{\method{} benefits from scaling both outer training rounds and inner exploration iterations.}
In this section, we further analyze two key scaling factors of \method{}:
(1) \textbf{the number of outer training rounds}. As shown in Figure~\ref{fig:more-analysis}(a), increasing the number of outer training rounds consistently improves performance on \textsc{WebVoyager}, yielding a total gain of 4.25\% from one round to three rounds.
(2) \textbf{the number of inner progressive exploration iterations}. As shown in Figure~\ref{fig:more-analysis}(b), increasing the number of inner exploration iterations also improves the overall success rate from 33.40\% at the initial rollout stage to 43.41\% after 15 stages.
Overall, \method{} shows favorable scaling behavior along both the policy-internalization and exploration-expansion dimensions: more outer rounds allow the policy to continually internalize task-completing skill chains discovered during exploration, while more inner iterations expand the reachable task frontier before each policy update.

\textbf{\method{} enables a self-evolving training loop.}
We investigate whether \method{} can replace the external memory manager with a self-managed one by fine-tuning a memory-management model from the same backbone as the policy model on pre-collected training data. As shown in Figure~\ref{fig:more-analysis}(c), self-managed memory achieves a final improvement comparable to the external-model setting, suggesting that \method{} can reduce reliance on external models and form a closed loop of self-evolution.

\crblue{\textbf{\method{} remains effective without auxiliary information.}
\method{} uses the auxiliary signal \(c\) in skill discovery and orchestration to align local-progress assessment with the standard task evaluator and provide limited evidence of progress toward task completion. A natural question is whether its performance gains depend on this additional information. To examine this, Table~\ref{tab:cr-webarena-controls} compares SCOPE with and without \(c\) on WebArena using Llama3.1-8B. Removing \(c\) from all stages largely preserves the \textsc{OPTI} reference's success rate (53.6\% with \(c\) versus 52.7\% without it). These results indicate that the auxiliary information serves as a lightweight signal for the memory-management model and contributes only marginally to final performance.}

\textbf{All components of \method{} are necessary for effective progressive learning.}
As shown in Figure~\ref{fig:more-analysis}(d), we evaluate three ablated variants: w/o Skill Discovery replaces semantic skill discovery with fixed-length trajectory segmentation every two steps; w/o Skill Orchestration replaces dynamic skill orchestration with a heuristic that always selects the deepest skill chain; and w/o Skill Expansion removes failure-experience-based frontier intervention, letting the policy continue self-exploration from the composed skill chain with a random temperature (tmp=1.0). Removing any component reduces performance; when all three are removed, the method degenerates to \textsc{RFT} and drops further, indicating that these modules play complementary roles in stable progressive learning.

\begin{table}[!t]
\centering
\begin{minipage}{\linewidth}
\centering
\makeatletter
\def\@captype{table}
\makeatother
\caption{\crblue{Auxiliary-information ablation on WebArena with Llama3.1-8B.}}
\vspace{-1mm}
\label{tab:cr-webarena-controls}
\begingroup
\fontsize{8.5}{10}\selectfont
\setlength{\tabcolsep}{6pt}
\begin{tabular}{lccccc}
\toprule
\textbf{Method} & \textbf{Total SR} & \textbf{Reddit} & \textbf{GitLab} & \textbf{Shopping} & \textbf{CMS} \\
\midrule
\textsc{WebRL} & 45.8 & 63.2 & 46.7 & 31.1 & 54.3 \\
\midrule
\rowcolor{lightbluebg}
\method{} w/ \(c\) & \textbf{53.6} & \textbf{84.2} & \textbf{61.4} & 40.0 & 51.4 \\
\rowcolor{lightbluebg}
\method{} w/o \(c\) & 52.7 & 68.4 & 53.3 & 40.0 & \textbf{60.0} \\
\bottomrule
\end{tabular}
\endgroup
\end{minipage}

\par\vspace{1.5em}
\begin{minipage}{\linewidth}
\centering
\makeatletter
\def\@captype{table}
\makeatother
\caption{\crblue{Model calls during training data collection, decomposed by stage. RFT corresponds to naive sampling. The WebShop GRPO count is approximate.}}
\vspace{-1mm}
\label{tab:cr-stage-cost}
\begingroup
\fontsize{8.5}{10}\selectfont
\setlength{\tabcolsep}{4pt}
\begin{tabular}{clrrrrr}
\toprule
\textbf{Model} & \textbf{Method} & \textbf{Policy Rollout} & \textbf{Discovery} & \textbf{Orchestration} & \textbf{Expansion} & \textbf{Total} \\
\midrule
\multicolumn{7}{c}{\it WebArena} \\
\midrule
 & \textsc{RFT} & 24,618 & 0 & 0 & 0 & 24,618 \\
\rowcolor{lightbluebg}
\cellcolor{white}\multirow[c]{-2}{*}{Qwen2.5-3B} & \method{} & 20,359 & 2,107 & 1,569 & 1,034 & 25,069 \\
\modelgroupcdashline{1-7}
 & \textsc{RFT} & 23,250 & 0 & 0 & 0 & 23,250 \\
\rowcolor{lightbluebg}
\cellcolor{white}\multirow[c]{-2}{*}{Llama3.1-8B} & \method{} & 17,957 & 1,902 & 1,358 & 959 & \textbf{22,176} \\
\midrule
\multicolumn{7}{c}{\it WebShop} \\
\midrule
 & \textsc{RFT} & 168,621 & 0 & 0 & 0 & 168,621 \\
 & \textsc{GRPO} & $\approx$242,000 & 0 & 0 & 0 & $\approx$242,000 \\
\rowcolor{lightbluebg}
\cellcolor{white}\multirow[c]{-3}{*}{Qwen2.5-3B} & \method{} & 92,354 & 7,767 & 3,067 & 5,541 & \textbf{108,729} \\
\bottomrule
\end{tabular}
\endgroup
\end{minipage}

\end{table}

\textbf{\method{} improves the cost--benefit trade-off.}
Training agents in real environments incurs two costs: model calls and environment interactions. We therefore evaluate training efficiency on \textsc{WebArena} by measuring the number of model calls and environment interactions required per successful task discovered during exploration. Compared with naive sampling, \method{} introduces memory-management calls between rollout rounds. However, these calls help the agent reuse verified local skills and avoid repeated low-value exploration, reducing the cost of discovering successful tasks. As shown in Figure~\ref{fig:more-analysis}(e), \method{} improves cost--benefit trade-off across model scales, discovering more reward-positive samples with fewer environment interactions and model calls.

\crblue{\textbf{Fine-grained model-call cost analysis.} Table~\ref{tab:cr-stage-cost} reports total model calls at each stage. Overall, on WebArena, SCOPE has a model-call cost comparable to RFT/naive sampling, requiring 1.8\% more calls with Qwen and 4.6\% fewer with Llama. On WebShop, SCOPE uses 35.5\% fewer model calls than RFT and 55.1\% fewer than GRPO. These results show that SCOPE achieves better performance with comparable or fewer model calls.}

\section{Conclusion}
We presented \method{}, a progressive exploration framework for long-horizon Web agents. \method{} decomposes exploration into a short-to-long skill-acquisition process that discovers skills and failure experiences, composes them into task-completing skill chains, and expands outward from them. The resulting task-completing skill chains are then internalized into the policy through agentic on-policy distillation. Experiments across realistic and simulated Web environments show that \method{} consistently improves exploration dynamics and final task performance across model backbones.

\begingroup
\renewcommand{\acksection}{\section*{Acknowledgments}}
\begin{ack}
This work is supported in part by Huawei Innovation Research Programs. We gratefully acknowledge the support from MindSpore, CANN, and Ascend AI Processor used for this research.
\end{ack}
\endgroup

\bibliographystyle{plainnat3}
\bibliography{citation}

@article{Wang_2024,
   title={A survey on large language model based autonomous agents},
   volume={18},
   ISSN={2095-2236},
   url={http://dx.doi.org/10.1007/s11704-024-40231-1},
   DOI={10.1007/s11704-024-40231-1},
   number={6},
   journal={Frontiers of Computer Science},
   publisher={Springer Science and Business Media LLC},
   author={Wang, Lei and Ma, Chen and Feng, Xueyang and Zhang, Zeyu and Yang, Hao and Zhang, Jingsen and Chen, Zhiyuan and Tang, Jiakai and Chen, Xu and Lin, Yankai and Zhao, Wayne Xin and Wei, Zhewei and Wen, Jirong},
   year={2024},
   month=mar }

@misc{zhou2024webarenarealisticwebenvironment,
      title={WebArena: A Realistic Web Environment for Building Autonomous Agents}, 
      author={Shuyan Zhou and Frank F. Xu and Hao Zhu and Xuhui Zhou and Robert Lo and Abishek Sridhar and Xianyi Cheng and Tianyue Ou and Yonatan Bisk and Daniel Fried and Uri Alon and Graham Neubig},
      year={2024},
      eprint={2307.13854},
      archivePrefix={arXiv},
      primaryClass={cs.AI},
      url={https://arxiv.org/abs/2307.13854}, 
}

@misc{yuan2023scalingrelationshiplearningmathematical,
      title={Scaling Relationship on Learning Mathematical Reasoning with Large Language Models}, 
      author={Zheng Yuan and Hongyi Yuan and Chengpeng Li and Guanting Dong and Keming Lu and Chuanqi Tan and Chang Zhou and Jingren Zhou},
      year={2023},
      eprint={2308.01825},
      archivePrefix={arXiv},
      primaryClass={cs.CL},
      url={https://arxiv.org/abs/2308.01825}, 
}

@misc{peng2019advantageweightedregressionsimplescalable,
      title={Advantage-Weighted Regression: Simple and Scalable Off-Policy Reinforcement Learning}, 
      author={Xue Bin Peng and Aviral Kumar and Grace Zhang and Sergey Levine},
      year={2019},
      eprint={1910.00177},
      archivePrefix={arXiv},
      primaryClass={cs.LG},
      url={https://arxiv.org/abs/1910.00177}, 
}

@misc{bai2024digirltraininginthewilddevicecontrol,
      title={DigiRL: Training In-The-Wild Device-Control Agents with Autonomous Reinforcement Learning}, 
      author={Hao Bai and Yifei Zhou and Mert Cemri and Jiayi Pan and Alane Suhr and Sergey Levine and Aviral Kumar},
      year={2024},
      eprint={2406.11896},
      archivePrefix={arXiv},
      primaryClass={cs.LG},
      url={https://arxiv.org/abs/2406.11896}, 
}

@misc{he2024webvoyagerbuildingendtoendweb,
      title={WebVoyager: Building an End-to-End Web Agent with Large Multimodal Models}, 
      author={Hongliang He and Wenlin Yao and Kaixin Ma and Wenhao Yu and Yong Dai and Hongming Zhang and Zhenzhong Lan and Dong Yu},
      year={2024},
      eprint={2401.13919},
      archivePrefix={arXiv},
      primaryClass={cs.CL},
      url={https://arxiv.org/abs/2401.13919}, 
}

@inproceedings{feng-etal-2024-large,
    title = "Large Language Model-based Human-Agent Collaboration for Complex Task Solving",
    author = "Feng, Xueyang  and
      Chen, Zhi-Yuan  and
      Qin, Yujia  and
      Lin, Yankai  and
      Chen, Xu  and
      Liu, Zhiyuan  and
      Wen, Ji-Rong",
    editor = "Al-Onaizan, Yaser  and
      Bansal, Mohit  and
      Chen, Yun-Nung",
    booktitle = "Findings of the Association for Computational Linguistics: EMNLP 2024",
    month = nov,
    year = "2024",
    address = "Miami, Florida, USA",
    publisher = "Association for Computational Linguistics",
    url = "https://aclanthology.org/2024.findings-emnlp.72/",
    doi = "10.18653/v1/2024.findings-emnlp.72",
    pages = "1336--1357"
}

@misc{xu2025largereasoningmodelssurvey,
      title={Towards Large Reasoning Models: A Survey of Reinforced Reasoning with Large Language Models},
      author={Fengli Xu and Qianyue Hao and Zefang Zong and Jingwei Wang and Yunke Zhang and Jingyi Wang and Xiaochong Lan and Jiahui Gong and Tianjian Ouyang and Fanjin Meng and Chenyang Shao and Yuwei Yan and Qinglong Yang and Yiwen Song and Sijian Ren and Xinyuan Hu and Yu Li and Jie Feng and Chen Gao and Yong Li},
      year={2025},
      eprint={2501.09686},
      archivePrefix={arXiv},
      primaryClass={cs.AI},
      url={https://arxiv.org/abs/2501.09686},
}

@misc{agarwal2024onpolicydistillationlanguagemodels,
      title={On-Policy Distillation of Language Models: Learning from Self-Generated Mistakes},
      author={Rishabh Agarwal and Nino Vieillard and Yongchao Zhou and Piotr Stanczyk and Sabela Ramos and Matthieu Geist and Olivier Bachem},
      year={2024},
      eprint={2306.13649},
      archivePrefix={arXiv},
      primaryClass={cs.LG},
      url={https://arxiv.org/abs/2306.13649}
}

@misc{ross2011reductionimitationlearningstructured,
      title={A Reduction of Imitation Learning and Structured Prediction to No-Regret Online Learning},
      author={Stephane Ross and Geoffrey J. Gordon and J. Andrew Bagnell},
      year={2011},
      eprint={1011.0686},
      archivePrefix={arXiv},
      primaryClass={cs.LG},
      url={https://arxiv.org/abs/1011.0686}
}

@misc{yuan2023studentfriendlyknowledgedistillation,
      title={Student-friendly Knowledge Distillation},
      author={Mengyang Yuan and Bo Lang and Fengnan Quan},
      year={2023},
      eprint={2305.10893},
      archivePrefix={arXiv},
      primaryClass={cs.CV},
      url={https://arxiv.org/abs/2305.10893}
}

@misc{sodhi2024stepstackedllmpolicies,
      title={SteP: Stacked LLM Policies for Web Actions},
      author={Sodhi, Paloma and Branavan, S. R. K. and Artzi, Yoav and McDonald, Ryan},
      year={2024},
      eprint={2310.03720},
      archivePrefix={arXiv},
      primaryClass={cs.LG},
      url={https://arxiv.org/abs/2310.03720},
}

@inproceedings{huang-etal-2025-r2d2,
    title = "{R}2{D}2: Remembering, Replaying and Dynamic Decision Making with a Reflective Agentic Memory",
    author = "Huang, Tenghao  and
      Basu, Kinjal  and
      Abdelaziz, Ibrahim  and
      Kapanipathi, Pavan  and
      May, Jonathan  and
      Chen, Muhao",
    editor = "Che, Wanxiang  and
      Nabende, Joyce  and
      Shutova, Ekaterina  and
      Pilehvar, Mohammad Taher",
    booktitle = "Proceedings of the 63rd Annual Meeting of the Association for Computational Linguistics (Volume 1: Long Papers)",
    month = jul,
    year = "2025",
    address = "Vienna, Austria",
    publisher = "Association for Computational Linguistics",
    url = "https://aclanthology.org/2025.acl-long.1464/",
    doi = "10.18653/v1/2025.acl-long.1464",
    pages = "30318--30330",
    ISBN = "979-8-89176-251-0",
}

@misc{agashe2024agentsopenagentic,
      title={Agent S: An Open Agentic Framework that Uses Computers Like a Human},
      author={Agashe, Saaket and Han, Jiuzhou and Gan, Shuyu and Yang, Jiachen and Li, Ang and Wang, Xin Eric},
      year={2024},
      eprint={2410.08164},
      archivePrefix={arXiv},
      primaryClass={cs.AI},
      url={https://arxiv.org/abs/2410.08164},
}

@misc{qwen2025qwen25technicalreport,
      title={Qwen2.5 Technical Report},
      author={Qwen and Yang, An and Yang, Baosong and Zhang, Beichen and Hui, Binyuan and Zheng, Bo and Yu, Bowen and Li, Chengyuan and Liu, Dayiheng and Huang, Fei and Wei, Haoran and Lin, Huan and Yang, Jian and Tu, Jianhong and Zhang, Jianwei and Yang, Jianxin and Yang, Jiaxi and Zhou, Jingren and Lin, Junyang and Dang, Kai and Lu, Keming and Bao, Keqin and Yang, Kexin and Yu, Le and Li, Mei and Xue, Mingfeng and Zhang, Pei and Zhu, Qin and Men, Rui and Lin, Runji and Li, Tianhao and Tang, Tianyi and Xia, Tingyu and Ren, Xingzhang and Ren, Xuancheng and Fan, Yang and Su, Yang and Zhang, Yichang and Wan, Yu and Liu, Yuqiong and Cui, Zeyu and Zhang, Zhenru and Qiu, Zihan},
      year={2025},
      eprint={2412.15115},
      archivePrefix={arXiv},
      primaryClass={cs.CL},
      url={https://arxiv.org/abs/2412.15115},
}

@misc{grattafiori2024llama3herdmodels,
      title={The Llama 3 Herd of Models},
      author={Grattafiori, Aaron and Dubey, Abhimanyu and Jauhri, Abhinav and Pandey, Abhinav and Kadian, Abhishek and Al-Dahle, Ahmad and Letman, Aiesha and Mathur, Akhil and Schelten, Alan and Vaughan, Alex and Yang, Amy and Fan, Angela and Goyal, Anirudh and Hartshorn, Anthony and Yang, Aobo and Mitra, Archi and Sravankumar, Archie and Korenev, Artem and Hinsvark, Arthur and Rao, Arun and Zhang, Aston and Rodriguez, Aurelien and Gregerson, Austen and Spataru, Ava and Roziere, Baptiste and Biron, Bethany and Tang, Binh and Caucheteux, Charlotte and Nayak, Chaya and Bi, Chloe and Marra, Chris and McConnell, Chris and Keller, Christian and Touret, Christophe and Wu, Chunyang and Wong, Corinne and Ferrer, Cristian Canton and Nikolaidis, Cyrus and Allonsius, Damien and Song, Daniel and Pintz, Danielle and Livshits, Danny and Wyatt, Danny and Esiobu, David and Choudhary, Dhruv and Mahajan, Dhruv and Garcia-Olano, Diego and Perino, Diego and Hupkes, Dieuwke and Lakomkin, Egor and AlBadawy, Ehab and Lobanova, Elina and Dinan, Emily and Smith, Eric Michael and Radenovic, Filip and Guzman, Francisco and Zhang, Frank and Synnaeve, Gabriel and Lee, Gabrielle and Anderson, Georgia Lewis and Thattai, Govind and Nail, Graeme and Mialon, Gregoire and Pang, Guan and Cucurell, Guillem and Nguyen, Hailey and Korevaar, Hannah and Xu, Hu and Touvron, Hugo and Zarov, Iliyan and Ibarra, Imanol Arrieta and Kloumann, Isabel and Misra, Ishan and Evtimov, Ivan and Zhang, Jack and Copet, Jade and Lee, Jaewon and Geffert, Jan and Vranes, Jana and Park, Jason and Mahadeokar, Jay and Shah, Jeet and van der Linde, Jelmer and Billock, Jennifer and Hong, Jenny and Lee, Jenya and Fu, Jeremy and Chi, Jianfeng and Huang, Jianyu and Liu, Jiawen and Wang, Jie and Yu, Jiecao and Bitton, Joanna and Spisak, Joe and Park, Jongsoo and Rocca, Joseph and Johnstun, Joshua and Saxe, Joshua and Jia, Junteng and Alwala, Kalyan Vasuden and Prasad, Karthik and Upasani, Kartikeya and Plawiak, Kate and Li, Ke and Heafield, Kenneth and Stone, Kevin and El-Arini, Khalid and Iyer, Krithika and Malik, Kshitiz and Chiu, Kuenley and Bhalla, Kunal and Lakhotia, Kushal and Rantala-Yeary, Lauren and van der Maaten, Laurens and Chen, Lawrence and Tan, Liang and Jenkins, Liz and Martin, Louis and Madaan, Lovish and Malo, Lubo and Blecher, Lukas and Landzaat, Lukas and de Oliveira, Luke and Muzzi, Madeline and Pasupuleti, Mahesh and Singh, Mannat and Paluri, Manohar and Kardas, Marcin and Tsimpoukelli, Maria and Oldham, Mathew and Rita, Mathieu and Pavlova, Maya and Kambadur, Melanie and Lewis, Mike and Si, Min and Singh, Mitesh Kumar and Hassan, Mona and Goyal, Naman and Torabi, Narjes and Bashlykov, Nikolay and Bogoychev, Nikolay and Chatterji, Niladri and Zhang, Ning and Duchenne, Olivier and Alrassy, Patrick and Zhang, Pengchuan and Li, Pengwei and Vasic, Petar and Weng, Peter and Bhargava, Prajjwal and Dubal, Pratik and Krishnan, Praveen and Koura, Punit Singh and Xu, Puxin and He, Qing and Dong, Qingxiao and Srinivasan, Ragavan and Ganapathy, Raj and Calderer, Ramon and Cabral, Ricardo Silveira and Stojnic, Robert and Raileanu, Roberta and Maheswari, Rohan and Patel, Rohit and Girdhar, Rohit and Sauvestre, Romain and Polidoro, Ronnie and Sumbaly, Roshan and Taylor, Ross and Silva, Ruan and Hou, Rui and Wang, Rui and Hosseini, Saghar and Chennabasappa, Sahana and Singh, Sanjay and Bell, Sean and Kim, Seohyun Sonia and Edunov, Sergey and Nie, Shaoliang and Narang, Sharan and Raparthy, Sharath and Shen, Sheng and Wan, Shengye and Bhosale, Shruti and Zhang, Shun and Vandenhende, Simon and Batra, Soumya and Whitman, Spencer and Sootla, Sten and Collot, Stephane and Gururangan, Suchin and Borodinsky, Sydney and Herman, Tamar and Fowler, Tara and Sheasha, Tarek and Georgiou, Thomas and Scialom, Thomas and Speckbacher, Tobias and Mihaylov, Todor and Xiao, Tong and Karn, Ujjwal and Goswami, Vedanuj and Gupta, Vibhor and Ramanathan, Vignesh and Kerkez, Viktor and Gonguet, Vincent and Do, Virginie and Vogeti, Vish and Albiero, Vitor and Petrovic, Vladan and Chu, Weiwei and Xiong, Wenhan and Fu, Wenyin and Meers, Whitney and Martinet, Xavier and Wang, Xiaodong and Wang, Xiaofang and Tan, Xiaoqing Ellen and Xia, Xide and Xie, Xinfeng and Jia, Xuchao and Wang, Xuewei and Goldschlag, Yaelle and Gaur, Yashesh and Babaei, Yasmine and Wen, Yi and Song, Yiwen and Zhang, Yuchen and Li, Yue and Mao, Yuning and Coudert, Zacharie Delpierre and Yan, Zheng and Chen, Zhengxing and Papakipos, Zoe},
      year={2024},
      eprint={2407.21783},
      archivePrefix={arXiv},
      primaryClass={cs.AI},
      url={https://arxiv.org/abs/2407.21783},
}

@misc{deepseekai2025deepseekv32pushingfrontieropen,
      title={DeepSeek-V3.2: Pushing the Frontier of Open Large Language Models},
      author={DeepSeek-AI},
      year={2025},
      eprint={2512.02556},
      archivePrefix={arXiv},
      primaryClass={cs.CL},
      url={https://arxiv.org/abs/2512.02556},
}

@misc{zheng2024llamafactory,
      title={LLaMA-Factory: Unified Efficient Fine-Tuning of 100+ Language Models},
      author={Yaowei Zheng and Richong Zhang and Junhao Zhang and Yanhan Ye and Zheyan Luo and Yongqiang Ma},
      year={2024},
      eprint={2403.13372},
      archivePrefix={arXiv},
      primaryClass={cs.CL},
      url={https://arxiv.org/abs/2403.13372},
}

@inproceedings{kwon2023pagedattention,
  title={Efficient Memory Management for Large Language Model Serving with PagedAttention},
  author={Kwon, Woosuk and Li, Zhuohan and Zhuang, Siyuan and Sheng, Ying and Zheng, Lianmin and Yu, Cody Hao and Gonzalez, Joseph E. and Zhang, Hao and Stoica, Ion},
  booktitle={Proceedings of the 29th Symposium on Operating Systems Principles},
  pages={611--626},
  year={2023},
  doi={10.1145/3600006.3613165},
  url={https://doi.org/10.1145/3600006.3613165}
}

@inproceedings{yao2022webshop,
  title={WebShop: Towards Scalable Real-World Web Interaction with Grounded Language Agents},
  author={Yao, Shunyu and Chen, Howard and Yang, John and Narasimhan, Karthik},
  booktitle={ArXiv},
  year={preprint},
  url={https://arxiv.org/abs/2207.01206}
}

@inproceedings{he-etal-2025-openwebvoyager,
    title = "{O}pen{W}eb{V}oyager: Building Multimodal Web Agents via Iterative Real-World Exploration, Feedback and Optimization",
    author = "He, Hongliang  and
      Yao, Wenlin  and
      Ma, Kaixin  and
      Yu, Wenhao  and
      Zhang, Hongming  and
      Fang, Tianqing  and
      Lan, Zhenzhong  and
      Yu, Dong",
    editor = "Che, Wanxiang  and
      Nabende, Joyce  and
      Shutova, Ekaterina  and
      Pilehvar, Mohammad Taher",
    booktitle = "Proceedings of the 63rd Annual Meeting of the Association for Computational Linguistics (Volume 1: Long Papers)",
    month = jul,
    year = "2025",
    address = "Vienna, Austria",
    publisher = "Association for Computational Linguistics",
    url = "https://aclanthology.org/2025.acl-long.1336/",
    doi = "10.18653/v1/2025.acl-long.1336",
    pages = "27545--27564",
    ISBN = "979-8-89176-251-0"
}

@InProceedings{zhou2025pae,
  title = {Proposer-Agent-Evaluator ({PAE}): Autonomous Skill Discovery For Foundation Model Internet Agents},
  author = {Zhou, Yifei and Yang, Qianlan and Lin, Kaixiang and Bai, Min and Zhou, Xiong and Wang, Yu-Xiong and Levine, Sergey and Li, Li Erran},
  booktitle = {Proceedings of the 42nd International Conference on Machine Learning},
  pages = {79490--79528},
  year = {2025},
  volume = {267},
  series = {Proceedings of Machine Learning Research},
  publisher = {PMLR},
  url = {https://proceedings.mlr.press/v267/zhou25ah.html}
}

@inproceedings{yang2025sage,
  title = {Self-Guided Hierarchical Exploration for Generalist Foundation Model Web Agents},
  author = {Yang, Qianlan and Wang, Xiangjun and Perszyk, Danielle and Wang, Yu-Xiong},
  booktitle = {Advances in Neural Information Processing Systems},
  year = {2025},
  url = {https://openreview.net/forum?id=9twwDW60Bw}
}

@misc{gandhi2026gobrowsetrainingwebagents,
      title={Go-Browse: Training Web Agents with Structured Exploration}, 
      author={Apurva Gandhi and Graham Neubig},
      year={2026},
      eprint={2506.03533},
      archivePrefix={arXiv},
      primaryClass={cs.CL},
      url={https://arxiv.org/abs/2506.03533}
}

@misc{feng2025groupingrouppolicyoptimizationllm,
      title={Group-in-Group Policy Optimization for LLM Agent Training}, 
      author={Lang Feng and Zhenghai Xue and Tingcong Liu and Bo An},
      year={2025},
      eprint={2505.10978},
      archivePrefix={arXiv},
      primaryClass={cs.LG},
      url={https://arxiv.org/abs/2505.10978}
}

@misc{shao2024deepseekmathpushinglimitsmathematical,
      title={DeepSeekMath: Pushing the Limits of Mathematical Reasoning in Open Language Models}, 
      author={Zhihong Shao and Peiyi Wang and Qihao Zhu and Runxin Xu and Junxiao Song and Xiao Bi and Haowei Zhang and Mingchuan Zhang and Y. K. Li and Y. Wu and Daya Guo},
      year={2024},
      eprint={2402.03300},
      archivePrefix={arXiv},
      primaryClass={cs.CL},
      url={https://arxiv.org/abs/2402.03300}
}

@misc{song2026surveyonpolicydistillationlarge,
      title={A Survey of On-Policy Distillation for Large Language Models}, 
      author={Mingyang Song and Mao Zheng},
      year={2026},
      eprint={2604.00626},
      archivePrefix={arXiv},
      primaryClass={cs.LG},
      url={https://arxiv.org/abs/2604.00626}
}

@misc{zhao2026selfdistilledreasoneronpolicyselfdistillation,
      title={Self-Distilled Reasoner: On-Policy Self-Distillation for Large Language Models}, 
      author={Siyan Zhao and Zhihui Xie and Mengchen Liu and Jing Huang and Guan Pang and Feiyu Chen and Aditya Grover},
      year={2026},
      eprint={2601.18734},
      archivePrefix={arXiv},
      primaryClass={cs.LG},
      url={https://arxiv.org/abs/2601.18734}
}

@article{liu2024visualagentbench,
  title={VisualAgentBench: Towards Large Multimodal Models as Visual Foundation Agents},
  author={Liu, Xiao and Zhang, Tianjie and Gu, Yu and Iong, Iat Long and Xu, Yifan and Song, Xixuan and Zhang, Shudan and Lai, Hanyu and Liu, Xinyi and Zhao, Hanlin and others},
  journal={arXiv preprint arXiv:2408.06327},
  year={2024}
}

@inproceedings{lu2025intelligentgoexplore,
  title={Intelligent Go-Explore: Standing on the Shoulders of Giant Foundation Models},
  author={Lu, Cong and Hu, Shengran and Clune, Jeff},
  booktitle={International Conference on Learning Representations},
  year={2025},
  url={https://openreview.net/forum?id=apErWGzCAA}
}

@article{ecoffet2021firstreturn,
  title={First return, then explore},
  author={Ecoffet, Adrien and Huizinga, Joost and Lehman, Joel and Stanley, Kenneth O. and Clune, Jeff},
  journal={Nature},
  volume={590},
  number={7847},
  pages={580--586},
  year={2021},
  month={Feb},
  doi={10.1038/s41586-020-03157-9},
  url={https://doi.org/10.1038/s41586-020-03157-9}
}

@book{fitts1967human,
  title={Human Performance},
  author={Fitts, Paul M. and Posner, Michael I.},
  year={1967},
  publisher={Brooks/Cole}
}

@inproceedings{wei-etal-2025-webagent,
    title = "{W}eb{A}gent-R1: Training Web Agents via End-to-End Multi-Turn Reinforcement Learning",
    author = "Wei, Zhepei  and
      Yao, Wenlin  and
      Liu, Yao  and
      Zhang, Weizhi  and
      Lu, Qin  and
      Qiu, Liang  and
      Yu, Changlong  and
      Xu, Puyang  and
      Zhang, Chao  and
      Yin, Bing  and
      Yun, Hyokun  and
      Li, Lihong",
    editor = "Christodoulopoulos, Christos  and
      Chakraborty, Tanmoy  and
      Rose, Carolyn  and
      Peng, Violet",
    booktitle = "Proceedings of the 2025 Conference on Empirical Methods in Natural Language Processing",
    month = nov,
    year = "2025",
    address = "Suzhou, China",
    publisher = "Association for Computational Linguistics",
    url = "https://aclanthology.org/2025.emnlp-main.401/",
    doi = "10.18653/v1/2025.emnlp-main.401",
    pages = "7909--7928",
    ISBN = "979-8-89176-332-6"
}

@misc{qi2025webrltrainingllmweb,
      title={WebRL: Training LLM Web Agents via Self-Evolving Online Curriculum Reinforcement Learning}, 
      author={Zehan Qi and Xiao Liu and Iat Long Iong and Hanyu Lai and Xueqiao Sun and Wenyi Zhao and Yu Yang and Xinyue Yang and Jiadai Sun and Shuntian Yao and Tianjie Zhang and Wei Xu and Jie Tang and Yuxiao Dong},
      year={2025},
      eprint={2411.02337},
      archivePrefix={arXiv},
      primaryClass={cs.CL},
      url={https://arxiv.org/abs/2411.02337}, 
}
\appendix

\newpage

\section{Algorithm}
\label{app:algorithm}

\begin{algorithm}[h]
\caption{\method{} progressive skill exploration and distillation.}
\label{alg:scope}
\begin{algorithmic}[1]
\Require task distribution \(\mathcal{D}\), initial policy \(\pi_\theta\), inner exploration budget \(J\)
\Statex \textit{Notation:} \(\oplus\) denotes trajectory concatenation.
\State initialize skill memory \(\mathcal{B}_{\mathrm{skill}}\gets\emptyset\) and failure memory \(\mathcal{B}_{\mathrm{fail}}\gets\emptyset\)
\For{each outer training round}
    \State sample a task goal \(G\sim\mathcal{D}\) and obtain auxiliary task information \(c\)
    \State collect an initial rollout \(\tau_0\) from scratch using \(\pi_\theta\)
    \State initialize teacher-trajectory buffer \(\mathcal{H}\gets\emptyset\)
    \For{\(j=1,\dots,J\)}
        \State \((\xi_i,y_i)_{i=1}^{n}\gets \textsc{Discover}(G,c,\tau_{j-1})\)
        \State \(\mathcal{B}_{\mathrm{skill}}\gets \mathcal{B}_{\mathrm{skill}}\cup\{\xi_i\mid y_i=1\}\)
        \State \(\mathcal{B}_{\mathrm{fail}}\gets \mathcal{B}_{\mathrm{fail}}\cup\{\xi_i\mid y_i=0\}\)
        \State build candidate frontier set \(\mathcal{S}_{\mathrm{frontier}}\) from \(\mathcal{B}_{\mathrm{skill}}\)
        \State select \(s^\star\gets \textsc{Select}(G,c,\mathcal{S}_{\mathrm{frontier}})\)
        \State compose and execute a skill chain \(p^\star\) from the initial state to \(s^\star\)
        \State retrieve matched failures \(\mathcal{F}^\star\gets\{\xi\in\mathcal{B}_{\mathrm{fail}}\mid s_{\texttt{start}}(\xi)=s^\star\}\)
        \If{using \textsc{OPTI}}
            \State generate frontier action \(a^\star\gets \textsc{Intervene}(G,s^\star,\mathcal{F}^\star)\)
        \Else
            \State generate frontier action \(a^\star\sim\pi_\theta(\cdot\mid G,s^\star,\textsc{Summarize}(G,p^\star,\mathcal{F}^\star))\)
        \EndIf
        \State execute \(a^\star\) at \(s^\star\) and observe the next state \(s^+\)
        \State continue rollout from \(s^+\) using policy \(\pi_\theta\), obtaining suffix rollout \(\tau_j^{+}\)
        \State construct \(\tau_j \gets p^\star \oplus a^\star \oplus \tau_j^{+}\)
        \If{\(R(\tau_j)=1\)}
            \State add the task-completing trajectory \(\tau_j\) to \(\mathcal{H}\)
        \EndIf
    \EndFor
    \State update \(\pi_\theta\) by minimizing \(\mathcal{L}_{\mathrm{distill}}(\theta)\) on trajectories in \(\mathcal{H}\)
\EndFor
\State \Return updated policy \(\pi_\theta\), skill memory \(\mathcal{B}_{\mathrm{skill}}\), failure memory \(\mathcal{B}_{\mathrm{fail}}\)
\end{algorithmic}
\end{algorithm}

\section{Related Work}
\label{app:related-work}

\paragraph{LLM-Based Web Agents.}
Large language models have driven autonomous agents from text generation toward real interactive environments. As a representative interface through which humans interact with the digital world, the Web has become an important testbed for autonomous-agent research. Existing work has introduced a range of Web interaction benchmarks and agent frameworks, such as \textsc{WebArena}~\citep{zhou2024webarenarealisticwebenvironment}, \textsc{WebVoyager}~\citep{he2024webvoyagerbuildingendtoendweb}, and \textsc{WebShop}~\citep{yao2022webshop}, to evaluate whether agents can complete user-specified tasks in real or simulated websites. In these environments, agents need to access target URLs through a browser and perform browser actions based on text-based webpage observations, such as HTML or AXTree. Early Web-agent methods mainly rely on prompting strong LLMs, using task decomposition, trajectory planning, webpage-observation compression, memory, and tool use to guide task execution~\citep{sodhi2024stepstackedllmpolicies,huang-etal-2025-r2d2,agashe2024agentsopenagentic}. In contrast, we focus on training lightweight open-source Web agents.

\paragraph{Web-Agent Training.}
Recent studies have begun to explore how to train Web agents, especially by allowing agents to interact directly with environments, collect behavioral data through trial and error, and optimize their policies using reward signals provided by environmental feedback. OpenWebVoyager~\citep{he-etal-2025-openwebvoyager} and PAE~\citep{zhou2025pae} exploit the asymmetry that generating and verifying Web tasks can be easier than solving them, and continuously create new tasks to improve agent capabilities. WebRL~\citep{qi2025webrltrainingllmweb} and SAGE~\citep{yang2025sage} focus on the challenge of training-task difficulty by constructing dynamic curricula that generate tasks according to the model's weaknesses on the training set. To address settings where the environment is difficult to characterize through predefined task generation, Go-Browser~\citep{gandhi2026gobrowsetrainingwebagents} proposes a bottom-up approach that discovers and constructs tasks during exploration. WebAgent-R1~\citep{wei-etal-2025-webagent} and GiGPO~\citep{feng2025groupingrouppolicyoptimizationllm} further optimize Web agents with online reinforcement learning from environment interactions. However, these methods still largely rely on naive rollouts: given a user query, the agent samples complete trajectories end to end from the initial state. In long-horizon Web tasks, such training processes often suffer from low exploration efficiency, scarce successful trajectories, and repeated failures. In contrast, \method{} goes beyond this paradigm by decomposing long and difficult queries into reusable local skills during exploration, then progressively composing and expanding these skills to support effective and sustained exploration.

\paragraph{On-Policy Distillation.}
Distillation is a natural alternative to sparse-reward reinforcement learning, since teacher trajectories can provide supervision at each step. However, offline distillation often suffers from an off-policy issue: fixed teacher data can be mismatched with the interaction distribution of the current policy in dynamic Web environments, and step-level imitation becomes ineffective once the student policy fails to reach the states in the teacher trajectories. To mitigate this problem, prior work has proposed OPD~\citep{song2026surveyonpolicydistillationlarge} and OPSD~\citep{zhao2026selfdistilledreasoneronpolicyselfdistillation}, whose core idea is to provide teacher supervision for each next token under states generated by the student LLM itself, often through techniques such as reverse KL divergence.

\section{Distillation View of \method{}}
\label{app:distillation-view}

The key advantages of \method{} can also be understood from the perspective of on-policy distillation. Distillation is a natural alternative to sparse-reward reinforcement learning, since teacher trajectories can provide denser step-level supervision for policy learning. However, conventional offline distillation is limited in Web-agent settings: (1) it imitates all behaviors in complete teacher trajectories rather than focusing on the key decisions that determine task progress; (2) it suffers from off-policy distribution shift in dynamic Web environments, where the student policy may fail to reach the states covered by fixed teacher data; and (3) it often depends on the availability of a strong external teacher.

From this perspective, we propose agentic OPD. Unlike conventional OPD, which mainly focuses on token-level distribution matching, we emphasize on-policy learning at the level of agent-environment interaction steps: the states used for distillation are environment states that the student policy can actually reach. \method{} addresses the limitations above through three designs:
\begin{itemize}[leftmargin=*, itemsep=0pt, topsep=2pt]
\item \textit{Skill Discovery} decomposes trajectories into subtask units, thereby exposing critical local progress and failure decision points rather than imitating external expert guidance at every step.
\item \textit{Skill Orchestration} composes skill chains from behaviors produced by the policy model's own historical rollouts, helping preserve the on-policy nature of the distilled supervision.
\item \textit{Skill Expansion} constructs intervention signals from failure experiences, providing corrective actions for subsequent exploration. The teacher constructed from failure experiences can co-evolve with the policy model; furthermore, the teacher role can also be played by the policy model itself, enabling self-distillation and reducing reliance on a strong external teacher.
\end{itemize}

\section{Theoretical Details}
\label{app:theory}

\subsection{Proof of the Effective-Frontier-State Lower Bound}

\paragraph{Lemma 1.}
Let $\alpha$ denote the \emph{intrinsic constraint dimension} of a Web task, i.e., the number of key task constraints that must be progressively satisfied in order to complete the task. Let $\delta_k$ denote the probability that, when the current memory frontier is at depth $k-1$, the agent selects an effective frontier state from the candidate frontier states that enables expansion from depth $k-1$ to depth $k$. If, at any fixed depth level, the number of webpage functional patterns relevant to the current subtask progression is bounded, then there exists a constant $C>0$ such that
\[
\delta_k \ge \frac{1}{Ck^\alpha}.
\]

\begin{proof}
Task progression can be viewed as gradually satisfying $\alpha$ key constraint dimensions. By construction, successful entries that are functionally identical and progress-equivalent lead to the same canonical frontier state. Therefore, within any fixed functional pattern, redundant, cyclic, or equivalent rollouts do not increase the number of distinguishable candidate frontier states.

Moreover, at any fixed depth level, the number of webpage functional patterns relevant to the current subtask progression is bounded. Hence, whenever the frontier advances from depth $t-1$ to depth $t$, each constraint dimension can introduce at most a constant number of new distinguishable states. Accumulated up to frontier depth $k$, each constraint dimension therefore has at most $O(k)$ distinguishable states.

Since a candidate frontier state is jointly determined by these $\alpha$ constraint dimensions, the total number of candidate frontier states at depth $k$ is at most
\[
O(k^\alpha).
\]
Thus, there exists a constant $C>0$ such that
\[
|\mathcal{S}_{\mathrm{frontier}}^{(k)}| \le Ck^\alpha,
\]
where \(\mathcal{S}_{\mathrm{frontier}}^{(k)}\) denotes the set of candidate frontier states used to expand the frontier from depth \(k-1\) to depth \(k\).

Since depth \(k\) is defined as a valid next expansion target of the current frontier, the set \(\mathcal{S}_{\mathrm{frontier}}^{(k)}\) contains at least one effective frontier state that enables expansion from depth \(k-1\) to depth \(k\). Therefore,
\[
\delta_k \ge \frac{1}{|\mathcal{S}_{\mathrm{frontier}}^{(k)}|} \ge \frac{1}{Ck^\alpha}.
\]
\end{proof}

\begin{figure}[t]
    \centering
    \includegraphics[width=\textwidth]{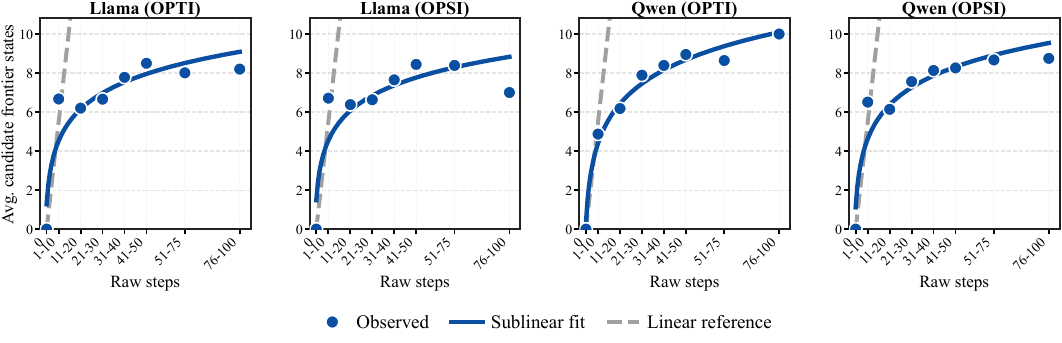}
    \caption{Growth of non-redundant candidate frontier states as raw exploration steps increase. Across model backbones and intervention variants, the number of candidate frontier states grows much more slowly than the linear reference \(y=x\), supporting the conservative nature of the counting bound.}
    \label{fig:frontier-growth}
\end{figure}

\paragraph{Remark on the Bound.}
Although the analysis above establishes a polynomial bound, our empirical results suggest that the practical growth pattern is often substantially better than this theoretical upper bound. Specifically, we examine how the size of the candidate frontier set \(|\mathcal{S}_{\mathrm{frontier}}^{(k)}|\) changes with the number of rollout steps. As shown in Figure~\ref{fig:frontier-growth}, across model backbones and intervention variants, the number of non-redundant candidate frontier states grows much more slowly than the linear reference \(y=x\), indicating a highly sublinear growth pattern in practice. This happens because, although a Web task may allow up to dozens of interaction steps, the number of functionally distinct webpage modules that can affect the next subtask is usually limited. This leads to much slower growth of frontier candidates in practice.

\subsection{Proof of Proposition 1}

\paragraph{Proposition 1 (Restated).}
Assume that making a locally correct frontier-expanding decision has probability lower bounded by $p \in (0,1]$. Under the polynomial lower bound on effective-frontier-state retrieval established in Lemma~1, the total number of exploration rollouts \(T_{\mathrm{total}}\) has polynomial variability in task depth; specifically,
\[
\mathrm{Std}(T_{\mathrm{total}})
=
O\!\left(\frac{L^{(1+2\alpha)/2}}{p}\right).
\]

\begin{proof}
Let \(T_k\) denote the number of exploration rollouts required to extend the frontier from depth \(k-1\) to depth \(k\) under self-composed progressive exploration, so that
\[
T_{\mathrm{total}}=\sum_{k=1}^{L}T_k.
\]
At depth $k$, the probability of selecting an effective frontier state is $\delta_k$, and conditioned on such a state, the probability of making a locally correct frontier-expanding decision is lower bounded by $p$. Hence the overall probability of extending the frontier at depth $k$ is
\[
q_k=p\delta_k,
\]
and under this abstraction,
\[
T_k \sim \mathrm{Geom}(q_k).
\]

For a geometric random variable with success probability $q_k$, we have
\[
\mathbb{E}[T_k]=\frac{1}{q_k},
\qquad
\mathrm{Var}(T_k)=\frac{1-q_k}{q_k^2}.
\]
Substituting $q_k=p\delta_k$ yields
\[
\mathbb{E}[T_k]=\frac{1}{p\delta_k},
\qquad
\mathrm{Var}(T_k)=\frac{1-p\delta_k}{p^2\delta_k^2}
=\Theta\!\left(\frac{1}{p^2\delta_k^2}\right).
\]

Assuming $\{T_k\}_{k=1}^L$ are independent under the stylized abstraction, it follows that
\[
\mathrm{Var}(T_{\mathrm{total}})
=
\sum_{k=1}^{L}\mathrm{Var}(T_k)
=
\Theta\!\left(\sum_{k=1}^{L}\frac{1}{p^2\delta_k^2}\right).
\]

By Lemma~1, there exists a constant $c>0$ such that
\[
\delta_k \ge c k^{-\alpha}.
\]
Hence
\[
\frac{1}{\delta_k^2}\le \frac{k^{2\alpha}}{c^2},
\]
and therefore
\[
\mathrm{Var}(T_{\mathrm{total}})
=
O\!\left(\frac{1}{p^2c^2}\sum_{k=1}^{L}k^{2\alpha}\right).
\]

Since
\[
\sum_{k=1}^{L}k^{2\alpha}=\Theta(L^{2\alpha+1}),
\]
we obtain
\[
\mathrm{Var}(T_{\mathrm{total}})
=
O\!\left(\frac{L^{2\alpha+1}}{p^2c^2}\right),
\qquad
\mathrm{Std}(T_{\mathrm{total}})
=
O\!\left(\frac{L^{(2\alpha+1)/2}}{p}\right),
\]
where the constant \(c\) from Lemma~1 is absorbed into the big-\(O\) notation.

Finally, under restart-from-scratch exploration, reaching depth \(k\) requires making correct decisions for all preceding \(k\) steps. Since each such decision succeeds with probability \(p\), the probability of reaching depth \(k\) is \(p^k\), so the number of rollouts required to reach depth \(k\) is geometric with mean \(\Theta(p^{-k})\), which grows exponentially in \(k\). In particular, completing a task of target depth \(L\) requires making all \(L\) critical decisions correctly, so the success probability becomes \(p^L\). Consequently, both the expected number of exploration rollouts required and its variability grow exponentially with task depth under restart-from-scratch exploration. In contrast, under self-composed progressive exploration, the variability remains polynomial, establishing the claimed advantage.
\end{proof}

\clearpage
\section{Case Study}
\label{app:case-study}

\begin{figure}[H]
    \centering
    \includegraphics[width=\textwidth]{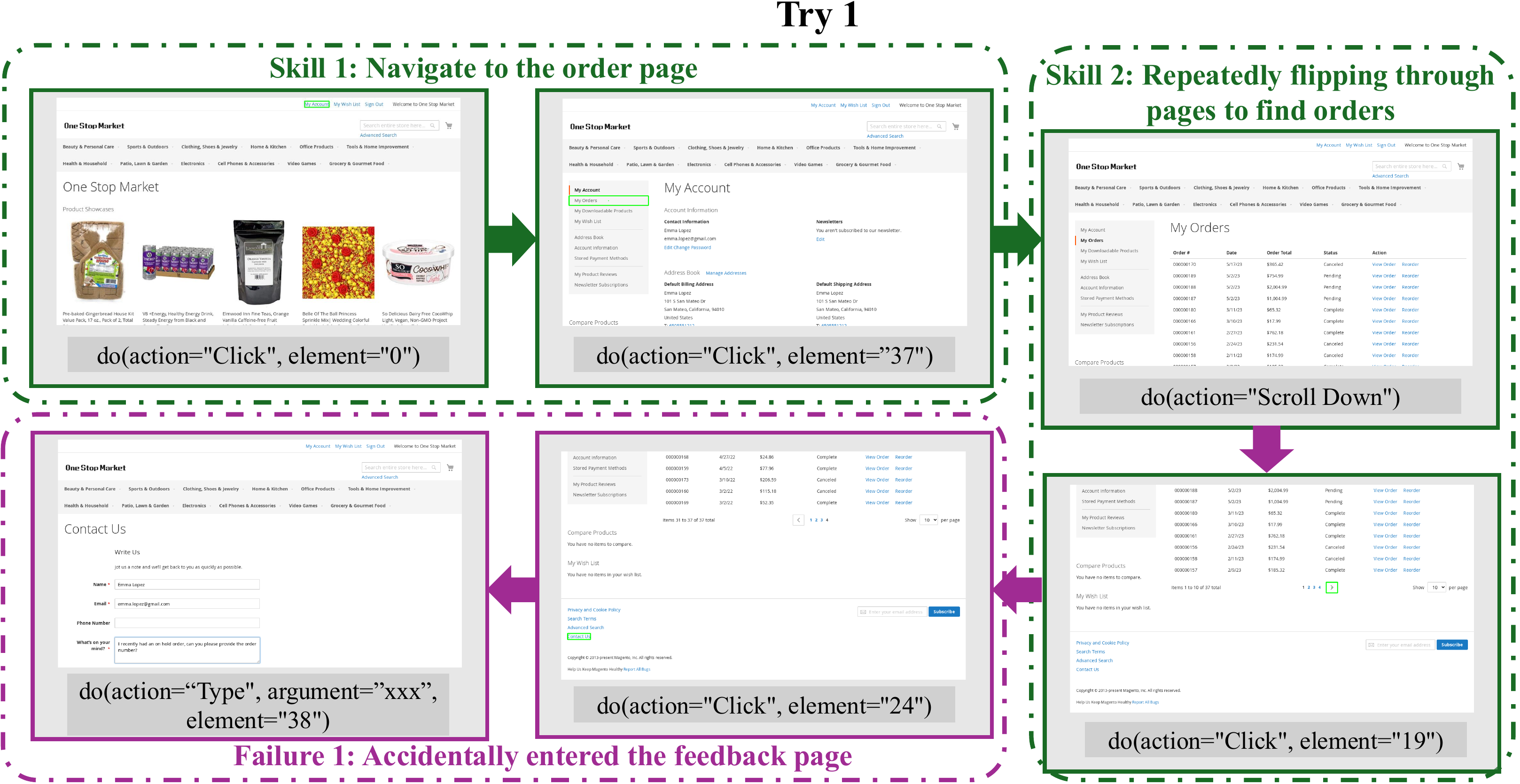}
    \vspace{0.6em}
    \includegraphics[width=\textwidth]{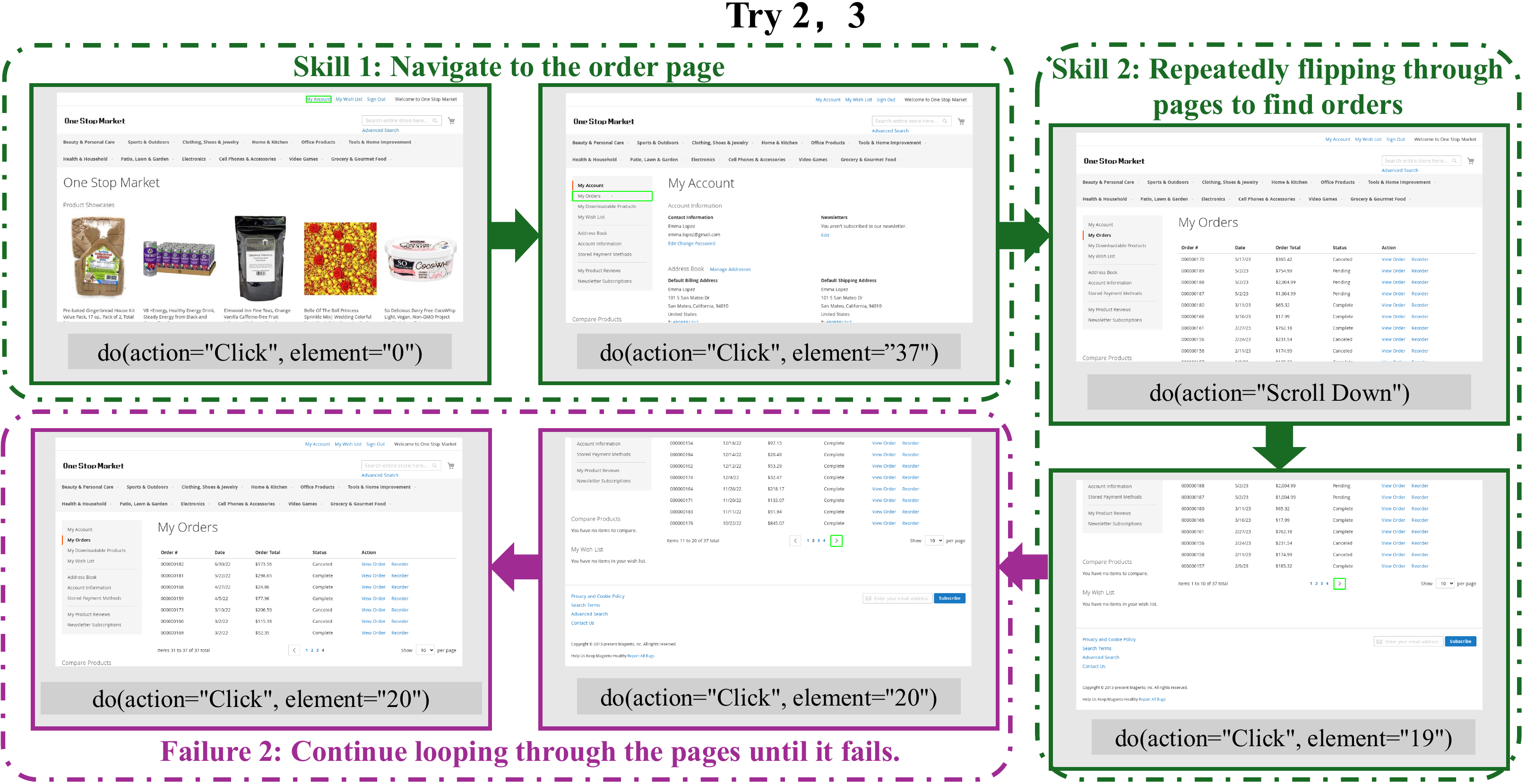}
    \caption{Early attempts in the case study. Top: the first attempt discovers reusable navigation and pagination skills, while recording a failure in which the agent accidentally enters the contact/feedback page. Bottom: the second and third attempts reuse the order-page navigation skill but expose a repeated failure pattern of flipping through order pages.}
    \label{fig:case-study-early}
\end{figure}

\begin{figure}[H]
    \centering
    \includegraphics[width=\textwidth]{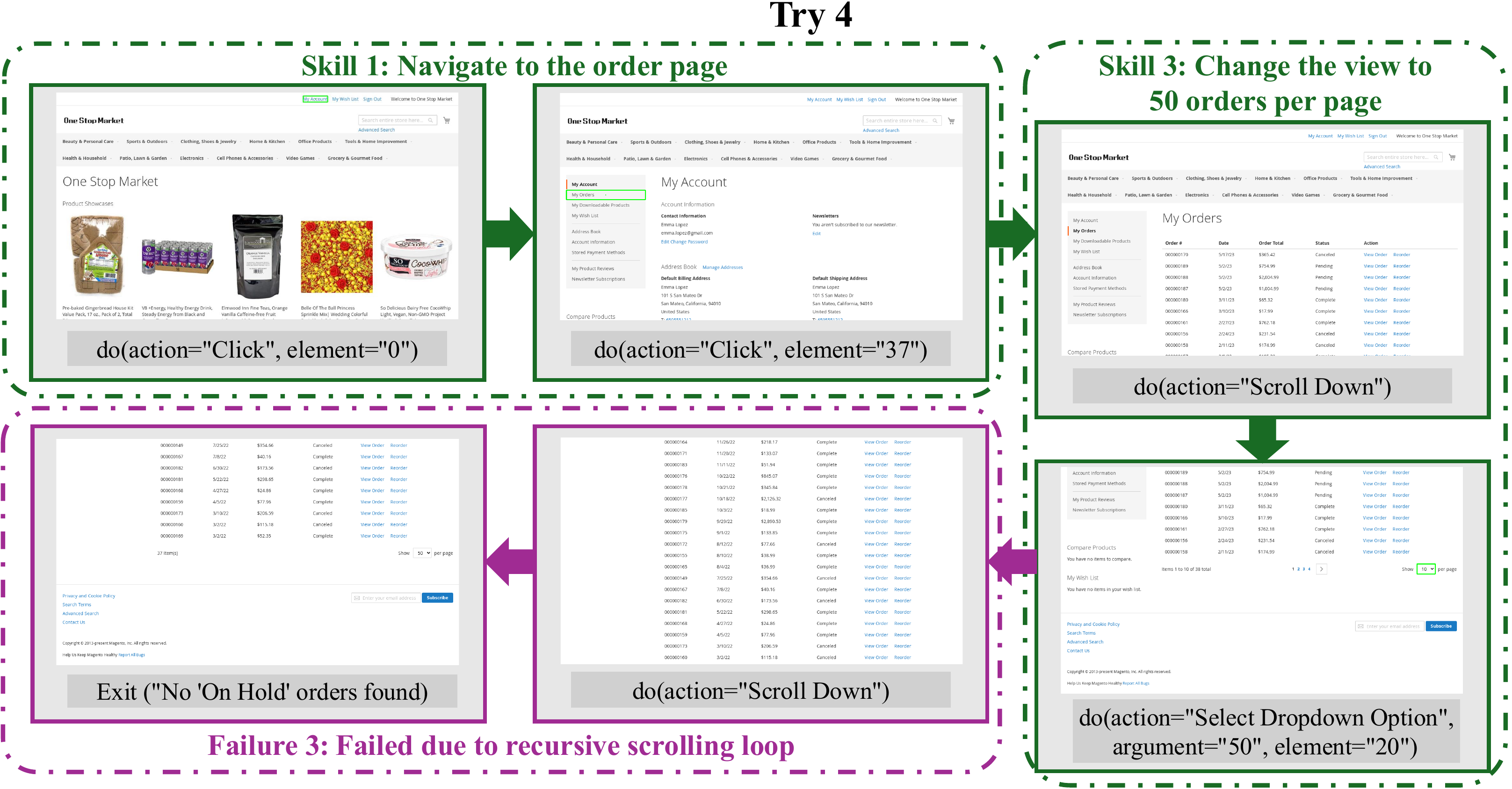}
    \vspace{0.6em}
    \includegraphics[width=\textwidth]{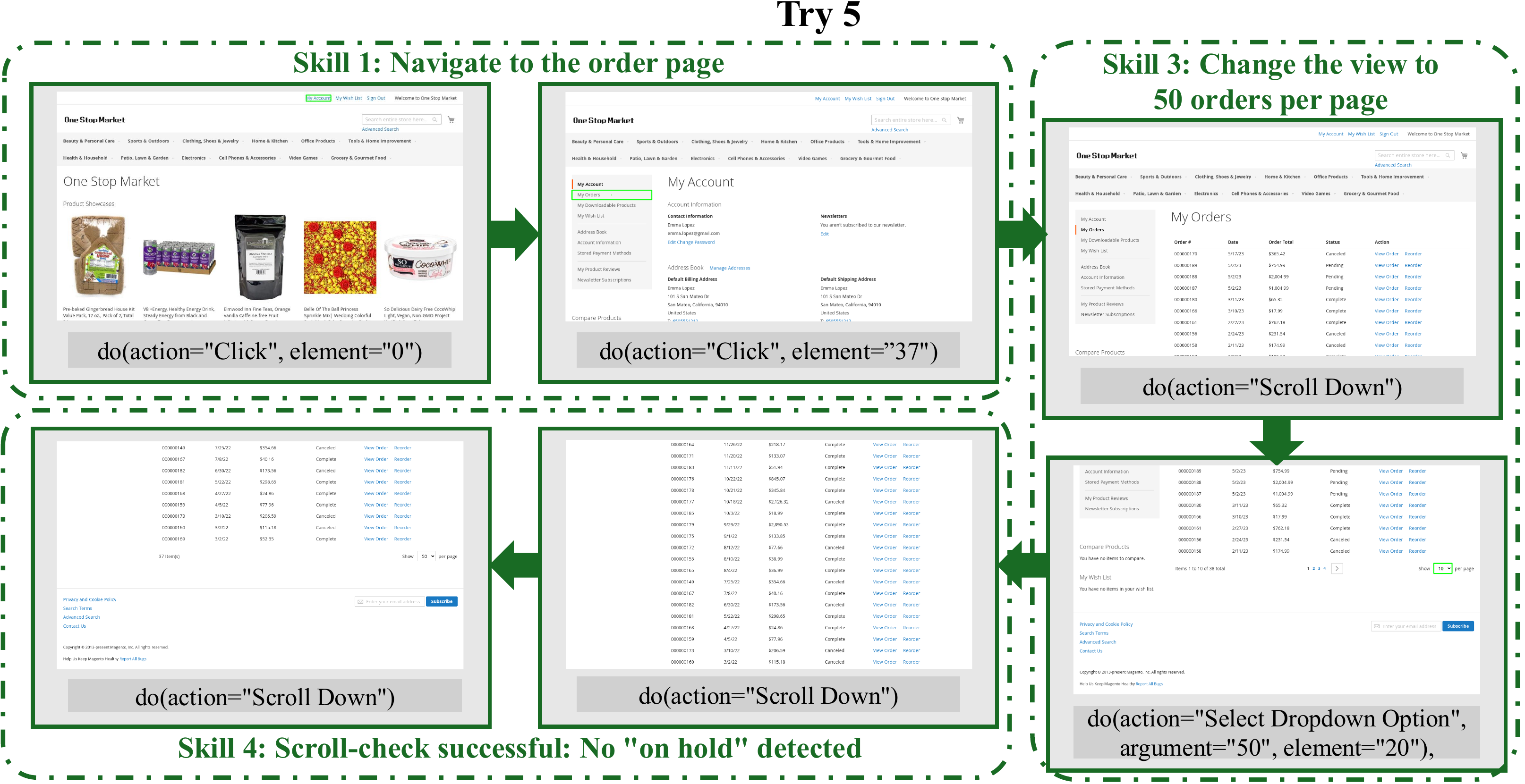}
    \caption{Later attempts in the case study. Top: in the fourth attempt, the agent changes the order list to show more entries per page, but still fails because the target information is not found after the view change. Bottom: in the final attempt, \method{} composes the navigation and view-changing skills into a skill chain and successfully completes the task on top of this chain.}
    \label{fig:case-study-later}
\end{figure}

Figures~\ref{fig:case-study-early} and~\ref{fig:case-study-later} present a short-to-long exploration example from a shopping task in WebArena. In the first attempt, the agent discovers skills for navigating from the store front page to the order page and for flipping through the order list, while also recording a failure pattern in which it accidentally enters the contact/feedback page. In the second and third attempts, the agent reuses the validated order-page navigation skill, but further exposes a repeated failure pattern of flipping through order pages. In the fourth attempt, the agent continues reusing existing skills and explores a new view-changing skill that displays more orders per page; however, this attempt still fails because the target information is not found after the view change. In the final attempt, \method{} composes a skill chain for entering the order page and changing the view, and successfully completes the task. This example illustrates the complementary roles of skill memory and failure memory: validated local skills preserve reusable progress, while failure records identify frontier states where subsequent exploration should be corrected.

\clearpage

\clearpage

\section{Additional Experimental Details}
\label{app:experimental-details}

\paragraph{Base Model Warm-Start Stage.}
Table~\ref{tab:appendix-data-sources} summarizes the supervised data sources used to initialize the policy models. Unless otherwise specified, we use the instruct versions of the corresponding backbone models as the base policy models. In the warm-up stage, we first apply supervised fine-tuning to the corresponding backbone model, and the data sources and scales used for fine-tuning are reported in Table~\ref{tab:appendix-data-sources}. The main exception is the Llama3.1-8B setting on \textsc{WebArena}, where we follow prior work~\citep{wei-etal-2025-webagent} and initialize from the \textsc{WebRL} checkpoint rather than directly from the instruct model. This setup matches prior WebArena training practice and provides a stronger Web-specific starting point for subsequent progressive exploration.

\paragraph{Auxiliary Information.}
During training, \method{} uses task-related auxiliary information \(c\) in \textsc{Discover} and \textsc{Select} to identify local progress and judge which frontier state is closer to the task goal. This section details how \(c\) is constructed. \(c\) is used to assist answer evaluation, such as the target URL and the expected states of webpage elements, but its form varies across datasets: (1) in \textbf{\textsc{WebArena}}, \(c\) includes the target URL and the states of task-specific target elements; (2) in \textbf{\textsc{WebShop}}, \(c\) consists of the constraint information of the target item; and (3) in \textbf{\textsc{WebVoyager}}, where such structured auxiliary information is unavailable, we instead use the evaluation reason produced by the LLM-as-judge for each rollout as the auxiliary signal. Overall, \(c\) is an auxiliary signal extracted from answer-evaluation information in the original dataset and related to task completion.

\begin{table}[h]
\centering
\small
\caption{Data sources and statistics for supervised policy initialization across benchmarks.}
\label{tab:appendix-data-sources}
\begin{adjustbox}{max width=\textwidth}
\begin{tabular}{lccc}
\toprule
\textbf{Benchmark} & \textbf{\textsc{SFT} data source} & \textbf{Trajectories} & \textbf{Steps} \\
\midrule
\textsc{WebArena} & VisualAgentBench~\citep{liu2024visualagentbench} & 1122 & 9460 \\
\textsc{WebVoyager} & DeepSeek sampling~\citep{deepseekai2025deepseekv32pushingfrontieropen} & 2987 & 15763 \\
\textsc{WebShop} & DeepSeek sampling~\citep{deepseekai2025deepseekv32pushingfrontieropen} & 2400 & 13337 \\
\bottomrule
\end{tabular}
\vspace{-1em}
\end{adjustbox}
\end{table}

\paragraph{Train/Test Splits and Rollout Budgets.}
Table~\ref{tab:appendix-splits-budgets} reports the train/test splits, task budgets, and \method{} iteration counts used in each benchmark. The total number of rollout attempts is matched between \textsc{RFT} and \method{}.

\begin{table}[h]
\centering
\normalsize
\caption{Train/test splits, task budgets, and \method{} iteration counts used in each benchmark.}
\label{tab:appendix-splits-budgets}
\setlength{\tabcolsep}{3.4pt}
\renewcommand{\arraystretch}{1.12}
\begin{adjustbox}{max width=\textwidth}
\begin{tabular}{lcccccc}
\toprule
\textbf{Benchmark} & \textbf{Reference setting} & \textbf{Train tasks} & \textbf{Test tasks} & \textbf{Max steps} & \textbf{Outer rounds} & \textbf{Inner iters.} \\
\midrule
\textsc{WebArena} & \textsc{WebRL}~\citep{qi2025webrltrainingllmweb} & 551 & 134 & 30 & 1 & 5 \\
\textsc{WebVoyager} & OpenWebVoyager~\citep{he-etal-2025-openwebvoyager} & 461 & 311 & 15 & 1/3 & 5 \\
\textsc{WebShop} & VERL-Agent~\citep{feng2025groupingrouppolicyoptimizationllm} & 6410 & 500 & 10 & 1 & 4 \\
\bottomrule
\end{tabular}
\vspace{-2em}
\end{adjustbox}
\renewcommand{\arraystretch}{1.0}
\end{table}

\paragraph{Auxiliary-Model Prompts.}
Figures~\ref{fig:discover-prompt},~\ref{fig:select-prompt}, and~\ref{fig:intervene-prompt} show the prompt templates for \textsc{Discover}, \textsc{Select}, and \textsc{Intervene}. \textsc{Discover} decomposes a trajectory into CMUs and identifies reusable progress and failure-causing branches, \textsc{Select} chooses the reusable prefix as restart anchor for the next rollout, and \textsc{Intervene} generates corrective guidance and the next action at the selected frontier state.

\begin{figure*}[t]
\begin{tcolorbox}[
  width=\textwidth,
  title=\small \textbf{\textsc{Discover} Prompt Template},
  fonttitle=\bfseries,
  arc=2mm,
  boxsep=5pt,
  top=5pt,
  bottom=5pt,
]
\small
You are a memory extractor for long-horizon Web or environment interaction tasks.\\

Your job is to analyze one trajectory and decompose it into multiple subtask completion segments (CMUs). Each CMU may consist of multiple atomic actions and should represent either reusable intermediate progress or an incorrect or failure-causing branch.\\

\textbf{You will be given:}\\
\texttt{- Task Goal}\\
\texttt{- Privileged Info}\\
\texttt{- Task Success}\\
\texttt{- Current Trajectory}\\

\textbf{met\_expectation Rules:}\\
\texttt{- Set to true} when the segment achieves meaningful intermediate progress toward the task goal.\\
\texttt{- Set to false} when the segment follows an incorrect branch, gets stuck, reaches an incorrect conclusion, or fails to finish correctly.\\
\texttt{- If task\_success=false}, the final CMU must be false and must include the failure-causing final action or failure region.\\

\textbf{Privileged-Mismatch Diagnosis:}\\
Use the privileged information to infer why the trajectory did not satisfy the task requirement. Do not use privileged information as an instruction for future actions; use it only to judge whether the observed trajectory satisfies the task.\\

\textbf{Required Output JSON Schema:}\\
Return JSON only. Do not include Markdown or any extra text.\\
\texttt{\{}\\
\texttt{\ \ "failure\_summary": "<complete trajectory recap + global path logic + one explicit failure reason>",}\\
\texttt{\ \ "cmus": [}\\
\texttt{\ \ \ \ \{}\\
\texttt{\ \ \ \ \ \ "start\_step\_id": <integer>,}\\
\texttt{\ \ \ \ \ \ "end\_step\_id": <integer>,}\\
\texttt{\ \ \ \ \ \ "skill\_summary": "<concrete subtask intent and operations>",}\\
\texttt{\ \ \ \ \ \ "status\_summary": "<precise ending state after this segment>",}\\
\texttt{\ \ \ \ \ \ "path\_logic": "From [Start Obs] to [Current Obs] via: [Action Sequence]",}\\
\texttt{\ \ \ \ \ \ "met\_expectation": <true|false>}\\
\texttt{\ \ \ \ \}}\\
\texttt{\ \ ]}\\
\texttt{\}}\\

\textbf{Hard Constraints:}\\
\texttt{- cmus} must be time-ordered, non-overlapping, and cover the full trajectory.\\
\texttt{- The first CMU} is a special root no-op segment with \texttt{start\_step\_id=-1} and \texttt{end\_step\_id=-1}.\\
\texttt{- ...}\\

\textbf{Input Context:}\\
\texttt{- Task Goal: \{\{goal\}\}}\\
\texttt{- Privileged Info: \{\{privileged\_info\}\}}\\
\texttt{- Task Success: \{\{task\_success\}\}}\\
\texttt{- Current Trajectory: \{\{trajectory\_data\}\}}
\end{tcolorbox}
\caption{Prompt template for \textsc{Discover}.}
\label{fig:discover-prompt}
\end{figure*}

\begin{figure*}[t]
\begin{tcolorbox}[
  width=\textwidth,
  title=\small \textbf{\textsc{Select} Prompt Template},
  fonttitle=\bfseries,
  arc=2mm,
  boxsep=5pt,
  top=5pt,
  bottom=5pt,
]
\small
Your task is to select the best reusable prefix as the restart anchor for the next rollout.\\

A prefix candidate is a reusable positive trajectory prefix from prefix memory.\\

\textbf{You are given:}\\
\texttt{- the final task goal (Global Goal),}\\
\texttt{- privileged information or oracle feedback,}\\
\texttt{- and a set of candidate prefixes with their associated intermediate states.}\\

Each prefix candidate may include:\\
\texttt{- a concise semantic description,}\\
\texttt{- the reached terminal state or boundary,}\\
\texttt{- select count.}\\

\textbf{Selection Rules:}\\
\texttt{- Prefer prefix candidates} whose terminal state is highly relevant to the global goal.\\
\texttt{- Prefer prefix candidates} that provide meaningful intermediate progress.\\
\texttt{- Avoid prefix candidates} that have been over-explored.\\
\texttt{- Avoid prefix candidates} whose downstream branches are repeatedly disproven, unless they still expose unexplored branches.\\
\texttt{- Do not select only by maximum depth.}\\

\textbf{Output Format:}\\
Return JSON only. No extra text. Return exactly one JSON object with exactly two keys.\\
\texttt{\{}\\
\texttt{\ \ "rationale": <string>,}\\
\texttt{\ \ "selected\_prefix\_id": <string>}\\
\texttt{\}}\\

\textbf{Input Context:}\\
\texttt{- Global Goal: \{\{goal\}\}}\\
\texttt{- Privileged Info: \{\{privileged\_info\}\}}\\
\texttt{- Prefix Memory (JSON list of prefix candidates): \{\{cmu\_trajectory\_json\}\}}
\end{tcolorbox}
\caption{Prompt template for \textsc{Select}.}
\label{fig:select-prompt}
\end{figure*}

\begin{figure*}[t]
\begin{tcolorbox}[
  width=\textwidth,
  title=\small \textbf{\textsc{Intervene} Prompt Template},
  fonttitle=\bfseries,
  arc=2mm,
  boxsep=5pt,
  top=5pt,
  bottom=5pt,
]
\small
You are a long-horizon environment interaction agent.\\

You are \textbf{NOT} starting from scratch. A reusable skill prefix has already been replayed, and you are now at the prefix frontier state.\\

Your task is to generate alternative action guidance and the next action based on prior failed trajectory reflections and the current step input.\\

\textbf{Your output format must strictly follow:}\\
\texttt{- <dataset-specific action space>}\\

Only output JSON in the following format:\\
\texttt{\{}\\
\texttt{\ \ "rational": "...",}\\
\texttt{\ \ "next\_action\_guidance": "...",}\\
\texttt{\ \ "next\_action": "..."}\\
\texttt{\}}\\

\textbf{Constraints:}\\
\texttt{1. Output exactly one JSON object.} Do not output markdown, code fences, or any extra explanation text.\\
\texttt{2. <dataset-specific constraints>}\\

\textbf{Task:}\\
\texttt{\{intent\}}\\

\textbf{Current step input (merged):}\\
\texttt{\{current\_step\_input\}}\\

\textbf{Matched failure memories for this prefix:}\\
\texttt{\{failure\_memories if failure\_memories else "(empty)"\}}\\

\textbf{Original failed behavior (from selected CMU actions):}\\
\texttt{\{original\_failed\_behavior if original\_failed\_behavior else "(empty)"\}}
\end{tcolorbox}
\caption{Prompt template for \textsc{Intervene}.}
\label{fig:intervene-prompt}
\end{figure*}

\section{Baseline and Benchmark Details.}
\subsection{Baseline Details}
\label{app:baseline-details}

\textbf{Prompting.} For prompting baselines, we follow the original setup of each dataset and directly prompt general-purpose models without fine-tuning. On \textsc{WebArena} and \textsc{WebVoyager}, we also report representative strong prompting models such as OAI-4o and DeepSeek-V3.2 where applicable.

\textbf{\textsc{SFT}.} \textsc{SFT} denotes standard supervised fine-tuning on offline trajectory data for each benchmark; the detailed data sources are summarized in Table~\ref{tab:appendix-data-sources}.

\textbf{\textsc{RFT}.} \textsc{RFT} denotes reject-sampling fine-tuning~\citep{yuan2023scalingrelationshiplearningmathematical}. Specifically, we first collect multiple rollout trajectories for each task, then filter for successful trajectories with \(R(\tau)=1\), and finally fine-tune the policy on these trajectories. We evaluate \textsc{RFT} on all three benchmarks. Unless otherwise specified, \textsc{RFT} and \method{} use the same total number of rollout attempts.

\textbf{\textsc{AWR and DigiRL}.} \textsc{AWR}~\citep{peng2019advantageweightedregressionsimplescalable} is an offline RL baseline that directly optimizes the policy model with an advantage-weighted regression loss. DigiRL~\citep{bai2024digirltraininginthewilddevicecontrol} builds on AWR by improving the advantage estimator and introducing an automatic curriculum. For further details, we refer readers to the \textsc{WebRL} paper~\citep{qi2025webrltrainingllmweb}. We directly take the results of these two baselines from the \textsc{WebRL} paper.

\textbf{\textsc{WebRL}.} \textsc{WebRL} is a representative and competitive Web-agent training baseline on \textsc{WebArena}~\citep{qi2025webrltrainingllmweb}. Its core idea is to use an external model to generate additional tasks similar to failed ones after Web-agent rollouts, thereby forming a curriculum-style learning process. For policy optimization, \textsc{WebRL} adopts an AWR-style off-policy RL learner.

\textbf{\textsc{GRPO}.} \textsc{GRPO} is a standard fully online RL method, and our implementation follows VERL-Agent~\citep{feng2025groupingrouppolicyoptimizationllm}. It samples a group of trajectories for each update and computes advantages within the group. However, because of its fully online nature, such methods are difficult to apply directly in non-stationary real Web environments with high interaction costs. We therefore include \textsc{GRPO} on \textsc{WebShop}~\citep{yao2022webshop} as a representative online RL baseline.

\subsection{Benchmark Details}
\label{app:benchmark-details}

Table~\ref{tab:benchmark-details} summarizes the benchmark websites, observation/action spaces, and evaluation protocols.

\begin{table*}[t]
\centering
\small
\caption{Benchmark details.}
\label{tab:benchmark-details}
\vspace{0.35em}
\setlength{\tabcolsep}{4pt}
\renewcommand{\arraystretch}{1.12}
\begin{adjustbox}{max width=\textwidth}
\begin{tabular}{>{\raggedright\arraybackslash}p{0.14\textwidth}>{\raggedright\arraybackslash}p{0.20\textwidth}>{\raggedright\arraybackslash}p{0.42\textwidth}>{\raggedright\arraybackslash}p{0.17\textwidth}}
\toprule
\textbf{Benchmark} & \textbf{Websites} & \textbf{Observation and action space} & \textbf{Evaluation} \\
\midrule
\textsc{WebArena}
& Reddit, Shopping, CMS, and GitLab.
& HTML-based observations. Actions: \texttt{Click}, \texttt{Right Click}, \texttt{Type}, \texttt{Search}, \texttt{Hover}, \texttt{Scroll Up}, \texttt{Scroll Down}, \texttt{Press Enter}, \texttt{Switch Tab}, \texttt{Select Dropdown Option}, \texttt{Wait}, \texttt{exit(message)}, \texttt{go\_backward()}, and \texttt{go\_forward()}.
& Rule matching + LLM-as-judge. \\
\midrule
\textsc{WebVoyager}
& Amazon, Apple, ArXiv, BBC, Cambridge, GitHub, HFace, and Wolfram.
& AXTree-based observations. Actions: \texttt{Click [id]}, \texttt{Type [id]; text}, \texttt{Scroll [id/WINDOW]; up/down}, \texttt{GoBack}, \texttt{Google}, \texttt{Wait}, and \texttt{ANSWER; text}.
& LLM-as-judge. \\
\midrule
\textsc{WebShop}
& A simulated shopping website.
& Actions: \texttt{search[keywords]} and \texttt{click[value]}.
& Rule matching. \\
\bottomrule
\end{tabular}
\vspace{-2em}
\end{adjustbox}
\renewcommand{\arraystretch}{1.0}
\end{table*}

\FloatBarrier
\begingroup
\section{Repeated WebVoyager Evaluation after Website Updates}
\label{app:cr-webvoyager}
During the rebuttal period, we evaluated each trained RFT and SCOPE model three times on six websites. Apple and arXiv underwent major website updates compared with the initial experiments, so we excluded them from this supplemental evaluation. The remaining websites are Amazon, BBC, Cambridge Dictionary, GitHub, Hugging Face, and Wolfram Alpha.

Avg@3 is the mean task success rate over the three evaluations. Pass@3 is the fraction of tasks solved at least once across those evaluations. We report the sample standard deviation of the three run-level overall success rates: RFT achieves 34.76\% $\pm$ 0.55 percentage points and SCOPE achieves 38.70\% $\pm$ 2.74 percentage points. SCOPE improves Avg@3 by 3.94 percentage points and Pass@3 by 8.53 percentage points.

\begin{table}[htbp]
\centering
\small
\caption{\crblue{Supplemental three-run WebVoyager evaluation on six websites (success rates, \%). Overall Avg@3 is reported as mean $\pm$ sample standard deviation in percentage points.}}
\label{tab:cr-webvoyager-repeat}
\begin{adjustbox}{width=0.99\textwidth, center}
\setlength{\tabcolsep}{2.5pt}
\begin{tabular}{llrrrrrrr}
\toprule
\textbf{Metric} & \textbf{Method} & \textbf{Total SR} & \textbf{Amazon} & \textbf{BBC} & \textbf{Cambridge} & \textbf{GitHub} & \textbf{HFace} & \textbf{Wolfram} \\
\midrule
Avg@3 & \textsc{RFT} & $34.76\pm0.55$ & 39.74 & 18.28 & 9.40 & 43.90 & 38.71 & 55.04 \\
\rowcolor{lightbluebg}
 & \method{} & $\mathbf{38.70}\pm2.74$ & 50.00 & 19.35 & 15.38 & 52.03 & 45.16 & 49.61 \\
\midrule
Pass@3 & \textsc{RFT} & 51.66 & 69.23 & 35.48 & 20.51 & 60.98 & 64.52 & 62.79 \\
\rowcolor{lightbluebg}
 & \method{} & \textbf{60.19} & 73.08 & 41.94 & 35.90 & 73.17 & 74.19 & 65.12 \\
\bottomrule
\end{tabular}
\end{adjustbox}
\end{table}
\FloatBarrier
\endgroup

\section{Limitations}
Our current implementation mainly considers text-based observations of Web tasks. Incorporating visual inputs, such as screenshots, may further improve performance because cues such as layout, position, and color provide important information for Web interaction. In addition, in real dynamic websites, the validity of learned skills may fluctuate as page states, interfaces, or backend responses change over time. Although \method{} uses robust skill execution to improve skill reusability, a small number of previously acquired skills may still become inapplicable under changed Web contexts.

\section{Broader Impacts}
This work conducts experiments primarily in realistic Web environments, where both the exploration process and the resulting stronger Web agents may occasionally produce unintended or undesirable actions on live websites. Although our experiments are conducted for research purposes, these risks highlight the importance of effective monitoring and control mechanisms when improving Web agents in real environments. In future work, developing reliable safeguards, task-level permissions, and human oversight for real-world Web-agent training and deployment will be important.

\clearpage

\newpage
\section*{NeurIPS Paper Checklist}

\begin{enumerate}

\item {\bf Claims}
    \item[] Question: Do the main claims made in the abstract and introduction accurately reflect the paper's contributions and scope?
    \item[] Answer: \answerYes{}
    \item[] Justification: The abstract and introduction state the main contributions and scope of \method{}. The theoretical discussion and experiments on WebArena, WebVoyager, and WebShop support the claimed improvements in long-horizon Web-agent training.
    \item[] Guidelines:
    \begin{itemize}
        \item The answer \answerNA{} means that the abstract and introduction do not include the claims made in the paper.
        \item The abstract and/or introduction should clearly state the claims made, including the contributions made in the paper and important assumptions and limitations. A \answerNo{} or \answerNA{} answer to this question will not be perceived well by the reviewers. 
        \item The claims made should match theoretical and experimental results, and reflect how much the results can be expected to generalize to other settings. 
        \item It is fine to include aspirational goals as motivation as long as it is clear that these goals are not attained by the paper. 
    \end{itemize}

\item {\bf Limitations}
    \item[] Question: Does the paper discuss the limitations of the work performed by the authors?
    \item[] Answer: \answerYes{}
    \item[] Justification: The paper includes a separate Limitations section. It discusses the current focus on text-based Web observations and the possible invalidation of learned skills under changing dynamic Web contexts.
    \item[] Guidelines:
    \begin{itemize}
        \item The answer \answerNA{} means that the paper has no limitation while the answer \answerNo{} means that the paper has limitations, but those are not discussed in the paper. 
        \item The authors are encouraged to create a separate ``Limitations'' section in their paper.
        \item The paper should point out any strong assumptions and how robust the results are to violations of these assumptions (e.g., independence assumptions, noiseless settings, model well-specification, asymptotic approximations only holding locally). The authors should reflect on how these assumptions might be violated in practice and what the implications would be.
        \item The authors should reflect on the scope of the claims made, e.g., if the approach was only tested on a few datasets or with a few runs. In general, empirical results often depend on implicit assumptions, which should be articulated.
        \item The authors should reflect on the factors that influence the performance of the approach. For example, a facial recognition algorithm may perform poorly when image resolution is low or images are taken in low lighting. Or a speech-to-text system might not be used reliably to provide closed captions for online lectures because it fails to handle technical jargon.
        \item The authors should discuss the computational efficiency of the proposed algorithms and how they scale with dataset size.
        \item If applicable, the authors should discuss possible limitations of their approach to address problems of privacy and fairness.
        \item While the authors might fear that complete honesty about limitations might be used by reviewers as grounds for rejection, a worse outcome might be that reviewers discover limitations that aren't acknowledged in the paper. The authors should use their best judgment and recognize that individual actions in favor of transparency play an important role in developing norms that preserve the integrity of the community. Reviewers will be specifically instructed to not penalize honesty concerning limitations.
    \end{itemize}

\item {\bf Theory assumptions and proofs}
    \item[] Question: For each theoretical result, does the paper provide the full set of assumptions and a complete (and correct) proof?
    \item[] Answer: \answerYes{}
    \item[] Justification: The theoretical assumptions and formal proofs are provided in Appendix~\ref{app:theory}. The appendix includes the proof of the effective-prefix lower bound and the proof of Proposition 1.
    \item[] Guidelines:
    \begin{itemize}
        \item The answer \answerNA{} means that the paper does not include theoretical results. 
        \item All the theorems, formulas, and proofs in the paper should be numbered and cross-referenced.
        \item All assumptions should be clearly stated or referenced in the statement of any theorems.
        \item The proofs can either appear in the main paper or the supplemental material, but if they appear in the supplemental material, the authors are encouraged to provide a short proof sketch to provide intuition. 
        \item Inversely, any informal proof provided in the core of the paper should be complemented by formal proofs provided in appendix or supplemental material.
        \item Theorems and Lemmas that the proof relies upon should be properly referenced. 
    \end{itemize}

    \item {\bf Experimental result reproducibility}
    \item[] Question: Does the paper fully disclose all the information needed to reproduce the main experimental results of the paper to the extent that it affects the main claims and/or conclusions of the paper (regardless of whether the code and data are provided or not)?
    \item[] Answer: \answerYes{}
    \item[] Justification: The Method and Experiments sections describe the proposed framework, benchmarks, baselines, evaluation metrics, and main experimental protocols. We also set the decoding temperature to 0 during inference to ensure deterministic and reproducible evaluation. Additional theoretical and qualitative details are provided in the appendix.
    \item[] Guidelines:
    \begin{itemize}
        \item The answer \answerNA{} means that the paper does not include experiments.
        \item If the paper includes experiments, a \answerNo{} answer to this question will not be perceived well by the reviewers: Making the paper reproducible is important, regardless of whether the code and data are provided or not.
        \item If the contribution is a dataset and\slash or model, the authors should describe the steps taken to make their results reproducible or verifiable. 
        \item Depending on the contribution, reproducibility can be accomplished in various ways. For example, if the contribution is a novel architecture, describing the architecture fully might suffice, or if the contribution is a specific model and empirical evaluation, it may be necessary to either make it possible for others to replicate the model with the same dataset, or provide access to the model. In general. releasing code and data is often one good way to accomplish this, but reproducibility can also be provided via detailed instructions for how to replicate the results, access to a hosted model (e.g., in the case of a large language model), releasing of a model checkpoint, or other means that are appropriate to the research performed.
        \item While NeurIPS does not require releasing code, the conference does require all submissions to provide some reasonable avenue for reproducibility, which may depend on the nature of the contribution. For example
        \begin{enumerate}
            \item If the contribution is primarily a new algorithm, the paper should make it clear how to reproduce that algorithm.
            \item If the contribution is primarily a new model architecture, the paper should describe the architecture clearly and fully.
            \item If the contribution is a new model (e.g., a large language model), then there should either be a way to access this model for reproducing the results or a way to reproduce the model (e.g., with an open-source dataset or instructions for how to construct the dataset).
            \item We recognize that reproducibility may be tricky in some cases, in which case authors are welcome to describe the particular way they provide for reproducibility. In the case of closed-source models, it may be that access to the model is limited in some way (e.g., to registered users), but it should be possible for other researchers to have some path to reproducing or verifying the results.
        \end{enumerate}
    \end{itemize}

\item {\bf Open access to data and code}
    \item[] Question: Does the paper provide open access to the data and code, with sufficient instructions to faithfully reproduce the main experimental results, as described in supplemental material?
    \item[] Answer: \answerYes{}
    \item[] Justification: We will provide anonymized code and data in the Appendix, together with instructions for reproducing the main experimental results.
    \item[] Guidelines:
    \begin{itemize}
        \item The answer \answerNA{} means that paper does not include experiments requiring code.
        \item Please see the NeurIPS code and data submission guidelines (\url{https://neurips.cc/public/guides/CodeSubmissionPolicy}) for more details.
        \item While we encourage the release of code and data, we understand that this might not be possible, so \answerNo{} is an acceptable answer. Papers cannot be rejected simply for not including code, unless this is central to the contribution (e.g., for a new open-source benchmark).
        \item The instructions should contain the exact command and environment needed to run to reproduce the results. See the NeurIPS code and data submission guidelines (\url{https://neurips.cc/public/guides/CodeSubmissionPolicy}) for more details.
        \item The authors should provide instructions on data access and preparation, including how to access the raw data, preprocessed data, intermediate data, and generated data, etc.
        \item The authors should provide scripts to reproduce all experimental results for the new proposed method and baselines. If only a subset of experiments are reproducible, they should state which ones are omitted from the script and why.
        \item At submission time, to preserve anonymity, the authors should release anonymized versions (if applicable).
        \item Providing as much information as possible in supplemental material (appended to the paper) is recommended, but including URLs to data and code is permitted.
    \end{itemize}

\item {\bf Experimental setting/details}
    \item[] Question: Does the paper specify all the training and test details (e.g., data splits, hyperparameters, how they were chosen, type of optimizer) necessary to understand the results?
    \item[] Answer: \answerYes{}
    \item[] Justification: The Experiments section specifies the evaluated benchmarks, model backbones, baselines, metrics, and held-out test setting used for the reported comparisons. Further implementation and analysis details are included in the Appendix.
    \item[] Guidelines:
    \begin{itemize}
        \item The answer \answerNA{} means that the paper does not include experiments.
        \item The experimental setting should be presented in the core of the paper to a level of detail that is necessary to appreciate the results and make sense of them.
        \item The full details can be provided either with the code, in appendix, or as supplemental material.
    \end{itemize}

\item {\bf Experiment statistical significance}
    \item[] Question: Does the paper report error bars suitably and correctly defined or other appropriate information about the statistical significance of the experiments?
    \item[] Answer: \answerNo{}
    \item[] Justification: \crblue{Most main comparisons use a single training run and evaluation. Appendix~\ref{app:cr-webvoyager} additionally reports three repeated WebVoyager evaluations, with sample standard deviations of run-level success rates.}
    \item[] Guidelines:
    \begin{itemize}
        \item The answer \answerNA{} means that the paper does not include experiments.
        \item The authors should answer \answerYes{} if the results are accompanied by error bars, confidence intervals, or statistical significance tests, at least for the experiments that support the main claims of the paper.
        \item The factors of variability that the error bars are capturing should be clearly stated (for example, train/test split, initialization, random drawing of some parameter, or overall run with given experimental conditions).
        \item The method for calculating the error bars should be explained (closed form formula, call to a library function, bootstrap, etc.)
        \item The assumptions made should be given (e.g., Normally distributed errors).
        \item It should be clear whether the error bar is the standard deviation or the standard error of the mean.
        \item It is OK to report 1-sigma error bars, but one should state it. The authors should preferably report a 2-sigma error bar than state that they have a 96\% CI, if the hypothesis of Normality of errors is not verified.
        \item For asymmetric distributions, the authors should be careful not to show in tables or figures symmetric error bars that would yield results that are out of range (e.g., negative error rates).
        \item If error bars are reported in tables or plots, the authors should explain in the text how they were calculated and reference the corresponding figures or tables in the text.
    \end{itemize}

\item {\bf Experiments compute resources}
    \item[] Question: For each experiment, does the paper provide sufficient information on the computer resources (type of compute workers, memory, time of execution) needed to reproduce the experiments?
    \item[] Answer: \answerYes{}
    \item[] Justification: The Experiments section includes a cost analysis in terms of environment interactions and model calls per unique successful task. Figure~\ref{fig:more-analysis}(e) reports the resulting training-exploration efficiency frontier.
    \item[] Guidelines:
    \begin{itemize}
        \item The answer \answerNA{} means that the paper does not include experiments.
        \item The paper should indicate the type of compute workers CPU or GPU, internal cluster, or cloud provider, including relevant memory and storage.
        \item The paper should provide the amount of compute required for each of the individual experimental runs as well as estimate the total compute. 
        \item The paper should disclose whether the full research project required more compute than the experiments reported in the paper (e.g., preliminary or failed experiments that didn't make it into the paper). 
    \end{itemize}
    
\item {\bf Code of ethics}
    \item[] Question: Does the research conducted in the paper conform, in every respect, with the NeurIPS Code of Ethics \url{https://neurips.cc/public/EthicsGuidelines}?
    \item[] Answer: \answerYes{}
    \item[] Justification: We have reviewed the NeurIPS Code of Ethics and believe the research conforms to it. The work evaluates Web agents on established benchmarks and does not involve human-subject experiments or the release of high-risk data.
    \item[] Guidelines:
    \begin{itemize}
        \item The answer \answerNA{} means that the authors have not reviewed the NeurIPS Code of Ethics.
        \item If the authors answer \answerNo, they should explain the special circumstances that require a deviation from the Code of Ethics.
        \item The authors should make sure to preserve anonymity (e.g., if there is a special consideration due to laws or regulations in their jurisdiction).
    \end{itemize}

\item {\bf Broader impacts}
    \item[] Question: Does the paper discuss both potential positive societal impacts and negative societal impacts of the work performed?
    \item[] Answer: \answerYes{}
    \item[] Justification: The Broader Impacts section discusses risks that can arise when Web-agent exploration and stronger trained agents operate in realistic Web environments. It also notes the importance of monitoring, control mechanisms, task-level permissions, and human oversight for real-world Web-agent training and deployment.
    \item[] Guidelines:
    \begin{itemize}
        \item The answer \answerNA{} means that there is no societal impact of the work performed.
        \item If the authors answer \answerNA{} or \answerNo, they should explain why their work has no societal impact or why the paper does not address societal impact.
        \item Examples of negative societal impacts include potential malicious or unintended uses (e.g., disinformation, generating fake profiles, surveillance), fairness considerations (e.g., deployment of technologies that could make decisions that unfairly impact specific groups), privacy considerations, and security considerations.
        \item The conference expects that many papers will be foundational research and not tied to particular applications, let alone deployments. However, if there is a direct path to any negative applications, the authors should point it out. For example, it is legitimate to point out that an improvement in the quality of generative models could be used to generate Deepfakes for disinformation. On the other hand, it is not needed to point out that a generic algorithm for optimizing neural networks could enable people to train models that generate Deepfakes faster.
        \item The authors should consider possible harms that could arise when the technology is being used as intended and functioning correctly, harms that could arise when the technology is being used as intended but gives incorrect results, and harms following from (intentional or unintentional) misuse of the technology.
        \item If there are negative societal impacts, the authors could also discuss possible mitigation strategies (e.g., gated release of models, providing defenses in addition to attacks, mechanisms for monitoring misuse, mechanisms to monitor how a system learns from feedback over time, improving the efficiency and accessibility of ML).
    \end{itemize}
    
\item {\bf Safeguards}
    \item[] Question: Does the paper describe safeguards that have been put in place for responsible release of data or models that have a high risk for misuse (e.g., pre-trained language models, image generators, or scraped datasets)?
    \item[] Answer: \answerNA{}
    \item[] Justification: The paper does not release high-risk pretrained models, scraped datasets, or other artifacts that require special responsible-release safeguards.
    \item[] Guidelines:
    \begin{itemize}
        \item The answer \answerNA{} means that the paper poses no such risks.
        \item Released models that have a high risk for misuse or dual-use should be released with necessary safeguards to allow for controlled use of the model, for example by requiring that users adhere to usage guidelines or restrictions to access the model or implementing safety filters. 
        \item Datasets that have been scraped from the Internet could pose safety risks. The authors should describe how they avoided releasing unsafe images.
        \item We recognize that providing effective safeguards is challenging, and many papers do not require this, but we encourage authors to take this into account and make a best faith effort.
    \end{itemize}

\item {\bf Licenses for existing assets}
    \item[] Question: Are the creators or original owners of assets (e.g., code, data, models), used in the paper, properly credited and are the license and terms of use explicitly mentioned and properly respected?
    \item[] Answer: \answerYes{}
    \item[] Justification: Existing benchmarks, models, and related methods used in the paper are properly cited, including WebArena, WebVoyager, and WebShop. We follow the corresponding benchmark and model terms of use and do not redistribute third-party assets beyond the intended experimental setting.
    \item[] Guidelines:
    \begin{itemize}
        \item The answer \answerNA{} means that the paper does not use existing assets.
        \item The authors should cite the original paper that produced the code package or dataset.
        \item The authors should state which version of the asset is used and, if possible, include a URL.
        \item The name of the license (e.g., CC-BY 4.0) should be included for each asset.
        \item For scraped data from a particular source (e.g., website), the copyright and terms of service of that source should be provided.
        \item If assets are released, the license, copyright information, and terms of use in the package should be provided. For popular datasets, \url{paperswithcode.com/datasets} has curated licenses for some datasets. Their licensing guide can help determine the license of a dataset.
        \item For existing datasets that are re-packaged, both the original license and the license of the derived asset (if it has changed) should be provided.
        \item If this information is not available online, the authors are encouraged to reach out to the asset's creators.
    \end{itemize}

\item {\bf New assets}
    \item[] Question: Are new assets introduced in the paper well documented and is the documentation provided alongside the assets?
    \item[] Answer: \answerNA{}
    \item[] Justification: The paper does not introduce a new dataset, benchmark, or model asset requiring separate asset documentation.
    \item[] Guidelines:
    \begin{itemize}
        \item The answer \answerNA{} means that the paper does not release new assets.
        \item Researchers should communicate the details of the dataset\slash code\slash model as part of their submissions via structured templates. This includes details about training, license, limitations, etc. 
        \item The paper should discuss whether and how consent was obtained from people whose asset is used.
        \item At submission time, remember to anonymize your assets (if applicable). You can either create an anonymized URL or include an anonymized zip file.
    \end{itemize}

\item {\bf Crowdsourcing and research with human subjects}
    \item[] Question: For crowdsourcing experiments and research with human subjects, does the paper include the full text of instructions given to participants and screenshots, if applicable, as well as details about compensation (if any)? 
    \item[] Answer: \answerNA{}
    \item[] Justification: The work does not involve crowdsourcing experiments or research with human subjects.
    \item[] Guidelines:
    \begin{itemize}
        \item The answer \answerNA{} means that the paper does not involve crowdsourcing nor research with human subjects.
        \item Including this information in the supplemental material is fine, but if the main contribution of the paper involves human subjects, then as much detail as possible should be included in the main paper. 
        \item According to the NeurIPS Code of Ethics, workers involved in data collection, curation, or other labor should be paid at least the minimum wage in the country of the data collector. 
    \end{itemize}

\item {\bf Institutional review board (IRB) approvals or equivalent for research with human subjects}
    \item[] Question: Does the paper describe potential risks incurred by study participants, whether such risks were disclosed to the subjects, and whether Institutional Review Board (IRB) approvals (or an equivalent approval/review based on the requirements of your country or institution) were obtained?
    \item[] Answer: \answerNA{}
    \item[] Justification: The work does not involve crowdsourcing experiments or research with human subjects, so IRB approval or equivalent review is not applicable.
    \item[] Guidelines:
    \begin{itemize}
        \item The answer \answerNA{} means that the paper does not involve crowdsourcing nor research with human subjects.
        \item Depending on the country in which research is conducted, IRB approval (or equivalent) may be required for any human subjects research. If you obtained IRB approval, you should clearly state this in the paper. 
        \item We recognize that the procedures for this may vary significantly between institutions and locations, and we expect authors to adhere to the NeurIPS Code of Ethics and the guidelines for their institution. 
        \item For initial submissions, do not include any information that would break anonymity (if applicable), such as the institution conducting the review.
    \end{itemize}

\item {\bf Declaration of LLM usage}
    \item[] Question: Does the paper describe the usage of LLMs if it is an important, original, or non-standard component of the core methods in this research? Note that if the LLM is used only for writing, editing, or formatting purposes and does \emph{not} impact the core methodology, scientific rigor, or originality of the research, declaration is not required.
    \item[] Answer: \answerYes{}
    \item[] Justification: LLMs are a core component of the research methodology. The Method and Experiments sections describe their use for Web-agent policies, progressive exploration, memory-guided intervention, and skill distillation.
    \item[] Guidelines:
    \begin{itemize}
        \item The answer \answerNA{} means that the core method development in this research does not involve LLMs as any important, original, or non-standard components.
        \item Please refer to our LLM policy in the NeurIPS handbook for what should or should not be described.
    \end{itemize}

\end{enumerate}

\end{document}